\documentclass[10pt]{article} 
\usepackage[preprint]{tmlr}

\usepackage{amsmath,amsfonts,bm}

\def\eqref#1{equation~\ref{#1}}

\def\1{\bm{1}}

\DeclareMathAlphabet{\mathsfit}{\encodingdefault}{\sfdefault}{m}{sl}
\SetMathAlphabet{\mathsfit}{bold}{\encodingdefault}{\sfdefault}{bx}{n}

\usepackage{hyperref}
\usepackage{url}
\usepackage{graphicx}  
\usepackage{booktabs}  
\usepackage{subcaption}
\usepackage{makecell}
\usepackage{multirow}
\usepackage{pifont}
\usepackage[table]{xcolor}
\usepackage[most]{tcolorbox}
\usepackage{amsthm}  
\usepackage{wrapfig}
\usepackage{listings}
\usepackage{algorithm}
\usepackage{algorithmic}
\newtheorem{theorem}{Theorem}

\newtheorem{definition}[theorem]{Definition}

\newtcbtheorem{observation}{Observation}{
  colback=pink!10,
  colframe=pink!60!black,
  fonttitle=\bfseries,
  coltitle=white,
  boxrule=0.8pt,
  arc=1mm,
  left=1mm, right=1mm, top=0.5mm, bottom=0.5mm,
  before skip=5pt, after skip=7pt,
  breakable  
}{obs}

\lstdefinelanguage{prompt}{
  basicstyle=\ttfamily\small,
  breaklines=true,
  breakindent=0pt,
  breakautoindent=false,
  frame=single,
  backgroundcolor=\color{pink!10},
  columns=fullflexible,
  showstringspaces=false,
  moredelim=**[is][\color{blue}]{<<}{>>}
}
\definecolor{aclblue}{RGB}{0,0,128}

\hypersetup{
    colorlinks=true,
    linkcolor=aclblue,
    citecolor=aclblue,
    urlcolor=aclblue
}

\title{NOSA: \underline{N}ative and \underline{O}ffloadable \underline{S}parse \underline{A}ttention}

\author{\textbf{Yuxiang Huang\textsuperscript{1}\thanks{\ \ indicates equal contribution.} , 
Pengjie Wang\textsuperscript{1}$^*$, 
Jicheng Han\textsuperscript{1}, 
Weilin Zhao\textsuperscript{1},
Zhou Su\textsuperscript{2},
Ao Sun\textsuperscript{1},}\\ 
\textbf{Hongya Lyu\textsuperscript{2}, 
Hengyu Zhao\textsuperscript{2},
Yudong Wang\textsuperscript{1,2},
Chaojun Xiao\textsuperscript{1}\thanks{\ \ Corresponding Authors.} , 
Xu Han\textsuperscript{1,3}$^\dag$, 
Zhiyuan Liu\textsuperscript{1,3}$^\dag$} \\
 \textsuperscript{1}{NLP Group, DCST, IAI, BNRIST, Tsinghua University, Beijing, China.}
\textsuperscript{2}{OpenBMB, China.}
\\
\textsuperscript{3}{Jiangsu Collaborative Innovation Center for Language Ability, Jiangsu Normal University, Xuzhou, China.}
\\
\texttt{\{huang-yx21,wpj24\}@mails.tsinghua.edu.cn, \{xcj,han-xu,liuzy\}@tsinghua.edu.cn}
}

\def\month{MM}  
\def\year{YYYY} 

\newcommand{\name}{NOSA}
\newcommand{\namei}{NOSI}

\begin{document}

\maketitle

\begin{abstract}

Decoding throughput improvements from larger inference batches are limited by GPU memory, which is largely consumed by the key-value (KV) cache.
Prior training-free KV cache offloading alleviates this by keeping redundant context on the CPU and fetching only a sparse subset for attention, but it often degrades long-generation quality due to training-inference mismatch on sparse patterns. 
Meanwhile, trainable sparse attention is incompatible with efficient offloading, as unconstrained KV accesses may force large CPU-to-GPU transfers and erase throughput gains.
To this end, we propose \name, a trainable sparse attention mechanism natively designed for KV cache offloading. \name~explicitly constrains the volume of CPU–GPU KV transfers, thereby achieving low communication overhead and high decoding throughput.
We further build \namei, a KV cache offloading inference system that fully unlocks \name's efficiency.
Empirical results on $\{1,3,8\}$B LLMs demonstrate that \name~outperforms KV cache offloading baselines on general, long-input, and long-generation tasks, while boosting decoding throughput by up to $5.04\times$, $1.92\times$, and $1.83\times$ over \textsc{FullAttn}, \textsc{InfLLMv2}, and \textsc{ShadowKV}, respectively.
We release our code at \url{https://github.com/thunlp/NOSA}\footnote{This paper has been accepted to EMNLP 2026 main conference.}.

\end{abstract}

\section{Introduction}

\begin{wrapfigure}{r}{0.5\textwidth}  
  \centering
  \vspace{-10pt}
  \includegraphics[width=\linewidth]{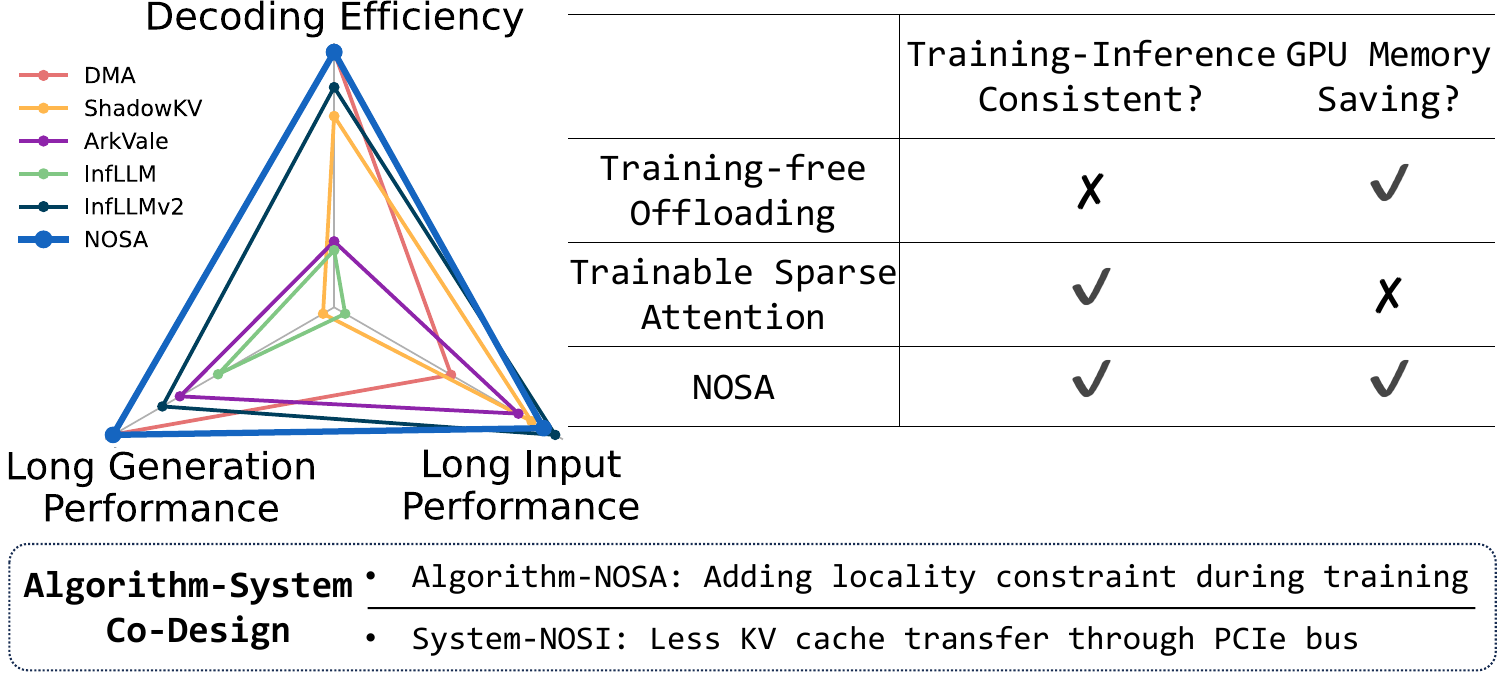}
  \vspace{-5pt}
  \caption{
    Overview of \name and \namei.
  }
  \vspace{-10pt}
  \label{fig:tri}
\end{wrapfigure}
Decoding efficiency of large language models (LLMs)~\citep{openai2025gpt, anthropic2025claude, deepseekai2024deepseekv32} is critical for advancing AI applications that require processing long-context inputs or outputs, such as LLM-based agents~\citep{qin2024tool, luo2025large} and solving complex reasoning tasks~\citep{guo2025deepseek, guo2025r}. However, this efficiency is heavily constrained by the $O(N)$-scale memory I/O for longer inputs.

To improve decoding efficiency, batched inference with KV-cache offloading is widely adopted~\citep{xiao2024infllm, sun2024shadowkv, yang2025lserve}. 
These methods keep most key-value (KV) cache units on the CPU and fetch only a small subset to the GPU for attention, reducing GPU memory footprint and enabling larger batches and higher decoding throughput. 
However, their training-free subset selection introduces a mismatch between training and inference, as the sparse patterns have not been seen during training, leading to substantial performance degradation (Figure~\ref{fig:tri}), particularly in long-generation scenarios where selection errors accumulate.

Meanwhile, trainable sparse attention methods~\citep{yuan2025native, lu2025moba, zhao2025infllm} can alleviate the training-inference gap, achieving near-lossless long-generation performance and outperforming prior attention-centric optimizations, including training-free sparse attention~\citep{li2025mminference, xu2025xattention} and KV-cache compression methods~\citep{zhang2023h2o, li2024snapkv}. However, these trainable methods do not reduce the size of the KV cache since any KV entry may be attended by a query, and thus cannot increase the maximum batch size under GPU memory constraints.
Therefore, \textbf{\textit{designing a trainable sparse attention mechanism that is natively offloadable}} is essential to address these challenges.

To this end, we propose \name~and \namei~to advance decoding acceleration with minimal performance degradation. We summarize our contributions as follows.

\underline{\textbf{Contribution 1:}} We first analyze the feasibility of building KV cache offloading systems based on trainable sparse attention methods. Using \textsc{InfLLMv2}~\citep{zhao2025infllm} as the representative of trainable sparse attention, we identify strong block-selection locality and theoretically analyze the resulting PCIe bottleneck. We show that the intrinsic locality of trainable sparse attention provides a natural basis for KV cache offloading; however, PCIe bandwidth limitations are difficult to overcome, making decoding become communication-bound.

\underline{\textbf{Contribution 2:}} Building on these insights, we introduce \name, a trainable sparse attention mechanism natively designed for offloading that explicitly limits KV cache transmission volume. \name~decomposes KV selection into query-aware and query-agnostic components, and applies an eviction policy over the query-agnostic selection to bound the number of KV blocks fetched from the CPU.
We train models integrated with \name~and conduct comprehensive evaluations on both short-context and long-context benchmarks. Experimental results show that \name~maintains good task performance and consistently outperforms all KV offloading baselines, especially on long-generation tasks.

\underline{\textbf{Contribution 3:}} We observe that vanilla offloading implementations are insufficient to fully expose the PCIe communication bottleneck. Accordingly, we develop \namei, an offloading system specifically tailored for \name. Through carefully designed system-level optimizations, \namei~achieves substantially higher decoding efficiency than existing baselines under various settings, delivering up to 5.04$\times$, 1.92$\times$, and 1.83$\times$ throughput improvement over \textsc{FullAttn}, \textsc{InfLLMv2}, and \textsc{ShadowKV}, respectively.

\begin{figure*}[t]
    \centering
    \begin{subfigure}{0.24\linewidth}
        \centering
        \includegraphics[width=\linewidth]{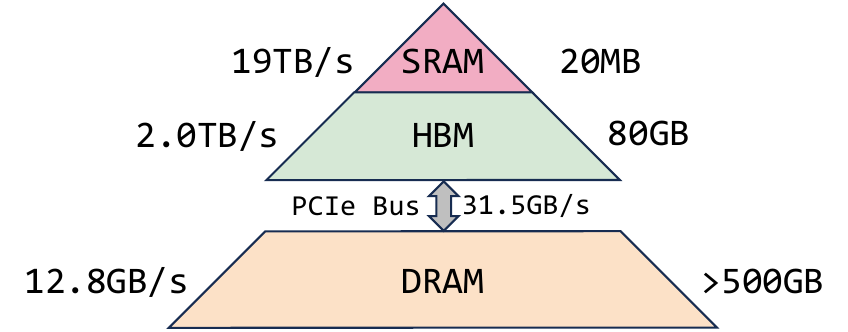}
        \vspace{3pt}
        \caption{Memory Hierarchy}
        \label{fig:cache-hit-rate-a}
    \end{subfigure}
    \begin{subfigure}{0.24\linewidth}
        \centering
        \includegraphics[width=\linewidth]{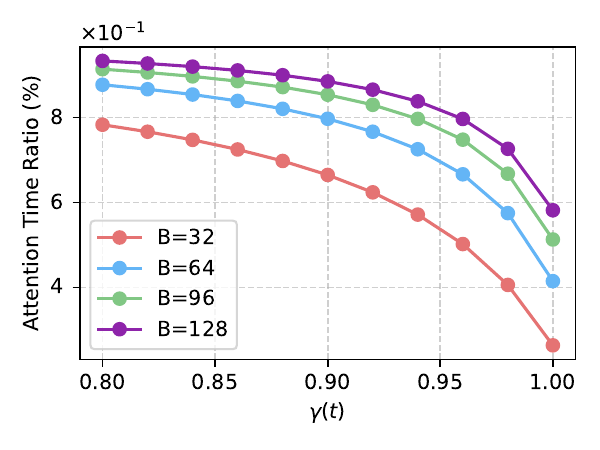}
        \vspace{-20pt}
        \caption{Attention Time Ratio}
        \label{fig:cache-hit-rate-b}
    \end{subfigure}
    \begin{subfigure}{0.24\linewidth}
        \centering
        \includegraphics[width=\linewidth]{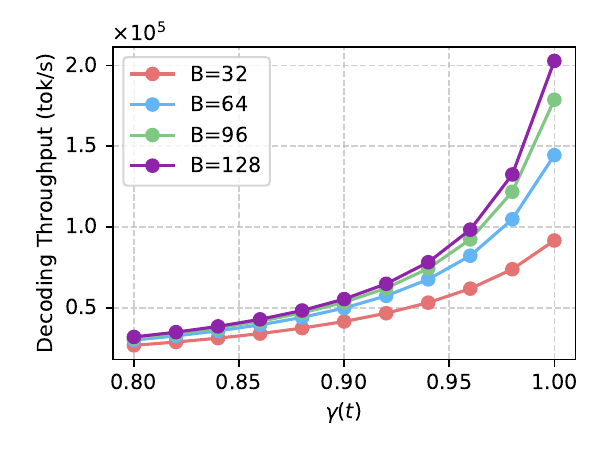}
        \vspace{-20pt}
        \caption{Theoretical Throughput}
        \label{fig:cache-hit-rate-c}
    \end{subfigure}
    \begin{subfigure}{0.24\linewidth}
        \centering
        \includegraphics[width=\linewidth]{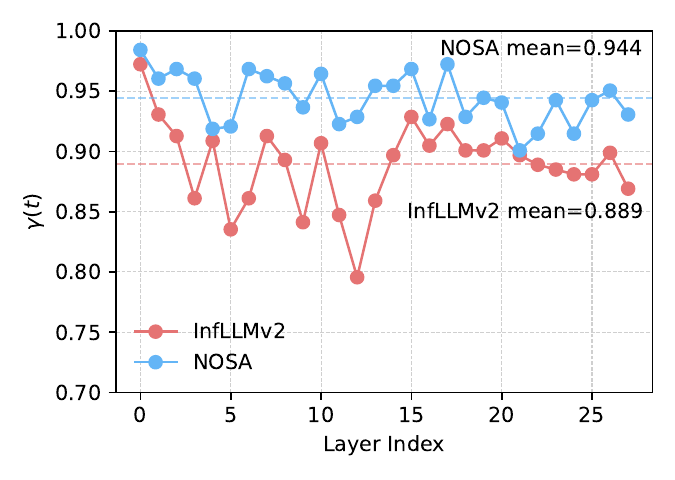}
        \vspace{-20pt}
        \caption{Locality $\gamma(t)$}
        \label{fig:infllmv2_locality}
    \end{subfigure}
    \caption{The memory hierarchy of NVIDIA A800-80GB, the ratio of attention mechanism among the complete inference time, the theoretical decoding throughput with respect to various cache hit rates, and \textsc{InfLLMv2}'s locality. ``B'' represents the batch size.}
    \vspace{-10pt} 
    \label{fig:cache-hit-rate}
\end{figure*}

\section{Background}

\subsection{Preliminaries}
\label{sec:preliminaries}
\label{sec:selection}

\textbf{Trainable Sparse Attention.} Transformer-based LLMs employ attention modules to capture cross-token relationships and long-range dependencies. 
The attention mechanism takes $\mathbf{H}\in\mathbb{R}^{n\times d}$ as input, where $n$ and $d$ denote the input length and model dimension, respectively. 
The query, key, and value matrices are then obtained through linear projections:
$
    \mathbf{Q} = \mathbf{H}\mathbf{W}_Q,~\mathbf{K} = \mathbf{H}\mathbf{W}_K,~ \mathbf{V} = \mathbf{H}\mathbf{W}_V.
$
The attention computation is then formulated as below, where $\mathbf{M}\in \{0, -\infty\}^{n\times n}$ is the attention mask.
\begin{align}
\label{eq:attn}
    \mathbf{P} = \mathbf{M} + \mathbf{Q}\mathbf{K}^\top,~ \mathbf{A} = \text{Softmax}(\mathbf{P}),~ \mathbf{O} = \mathbf{A}\mathbf{V}.
\end{align}
Trainable sparse attention methods commonly adopt a block-sparse attention pattern. 
Here, we briefly review the notations of \textsc{InfLLMv2}~\citep{zhao2025infllm}.
Before performing the attention computation, the key matrix is first partitioned into blocks of size $n_b$, and each block is compressed into a single representation using a compression function $f_c:\mathbb{R}^{n_b\times d}\rightarrow\mathbb{R}^{1\times d}$. 
This process can be formalized as follows:
$
    \mathbf{K}_c = f_c(\mathbf{K})\in \mathbb{R}^{\frac{n}{n_b}\times d}.
$
The selection score $\mathbf{S}_c^q = ({s}_{ij}^q)_{n\times \frac{n}{n_b}}$ between each token and all blocks is then computed as:
$
    {\mathbf{S}}_c^q = \mathbf{Q}\mathbf{K}_c^\top \in \mathbb{R}^{n\times \frac{n}{n_b}}.
$
Since \textsc{InfLLMv2} also incorporates attention sinks and sliding windows, we denote their lengths as $n_s$ and $n_w$, respectively. Then, the attention mask $\mathbf{M}=(m_{ij})_{n\times n}$ is computed as:
\begin{equation}
\begin{aligned}
&m_{ij} =
\begin{cases}
0, & \mathbb{I}_{\text{causal}} \land \left(\mathbb{I}_{\text{sink}} \lor \mathbb{I}_{\text{window}} \lor \mathbb{I}_{\text{topk}} \right); \\
-\infty, & \text{otherwise;}
\end{cases} \\
\mathbb{I}_{\text{causal}}=i \ge j;~ &\mathbb{I}_{\text{sink}}=j < n_s;~ \mathbb{I}_{\text{window}}=i - j < n_w;~\mathbb{I}_{\text{topk}}= s_{ij}^q \in
    \text{Top}_k\!\left(
        s^q_{i,\le \lfloor i/n_b \rfloor}
    \right)
\end{aligned}
\end{equation}
Finally, the attention is computed following Equation~\ref{eq:attn}.

\textbf{KV Cache Selection.} KV cache selection is widely used in sparse attention mechanisms, where a limited number of KV entries are accessed and computed. Mainstream approaches to KV cache selection can be broadly categorized into two types: \textit{query-aware selection}~\citep{yuan2025native, zhao2025infllm, li2024snapkv} and \textit{query-agnostic selection}~\citep{kim2024infinipot, huang2024locret, yao2024sirllm}. 

For query-aware selection, we first compute the selection score $\mathbf{S}^q = \mathbf{Q}\mathbf{K}^\top$, and then choose the top-$k$ positions, where $m_{ij} = 0$ if $s_{ij}^q \in \text{Top}_k(s_{i, \le i}^q)$. This utilizes information from the query to ensure the ability to recall distant information, but does not guarantee selection locality.

Query-agnostic selection is widely adopted in scenarios where the query tokens are not immediately available. To perform this selection, we first compute the importance score for each KV cache unit using a scoring function $s_i^e = f_e(\mathbf{h}_i)\in \mathbb{R}$, and then select the top-$k$ positions, where $m_{ij} = 0$ if $s_j^e \in \text{Top}_k(s_{\le i}^e)$. Query-agnostic selection does not guarantee the ability to recall and yields a fixed eviction pattern, which maintains strong locality and enables zero communication when applied to offloading systems.

\subsection{Observations}
\label{sec:observations}

We conduct an analytical study on the potential of applying trainable sparse attention mechanisms to offloading systems on typical compute architectures (as illustrated in Figure~\ref{fig:cache-hit-rate-a}). Based on this, we derive the following key observations.

{\small
\begin{observation}{Locality in Query-aware Selection}{locality}
Trainable sparse attention exhibits inherent locality in query-aware token/block selection across consecutive decoding steps.
\end{observation}}

We first analyze the locality of KV selection in a representative trainable sparse attention mechanism, \textsc{InfLLMv2}~\citep{zhao2025infllm}. Since KV cache offloading is primarily constrained by PCIe transmission overhead, higher locality allows caching and reusing KV caches selected in the previous decoding step for the current step, thereby reducing communication cost.

\begin{definition}[Locality]
Given an attention mask $\mathbf{M} = (m_{ij})_{n \times n}$, we define the set of selected tokens at the $t$-th step as $\Gamma(t) = \{ i : m_{it} = 0 \}$.
The locality at decoding step $t$ is defined as the fraction of tokens selected at step $t$ and also selected at step $t-1$: $\gamma(t) = \frac{|\Gamma(t) \cap \Gamma(t-1)|}{|\Gamma(t)|}$.
\end{definition}
We measure the locality $\gamma(t)$ for each transformer layer using an input sequence of 16K tokens, among which 4K tokens are selected. 
Figure~\ref{fig:infllmv2_locality} shows that, we have $\gamma(t)\!\ge\!0.8$ for most layers.
This strong locality implies that fewer than $20\%$ of the KV entries change per step and thus require communication, making KV cache offloading possible. Notably, this inherent locality emerges naturally from the trained sparse attention without any explicit constraints.

{\small
\begin{observation}{PCIe Communication Bound}{cache_hit}
    Query-aware offloading inference remains bounded by the communication latency of the PCIe bus.
\end{observation}}

Although Observation~\ref{obs:locality} indicates that high locality enables the potential for offloading by reusing previously selected tokens as cache, we find that a $\sim80\%$ locality is still insufficient to effectively mitigate PCIe communication overhead.
Figure~\ref{fig:cache-hit-rate-b} shows that more than $80\%$ of the decoding time is spent within the attention mechanism, suggesting that decoding throughput remains constrained by communication. 

To further analyze this effect, we simulate the theoretical inference latency and decoding throughput under varying cache hit rates (i.e., locality $\gamma(t)$), following the efficiency estimation protocol proposed in~\citet{yuan2024llm}. Figure~\ref{fig:cache-hit-rate-c} shows that increasing the cache hit rate significantly reduces the fraction of time spent on attention, thereby improving the theoretical decoding throughput.

\textbf{\underline{Takeaway:}} {Building on Observation~\ref{obs:locality} and~\ref{obs:cache_hit}, achieving high decoding throughput critically requires enforcing a training-time constraint that promotes higher cache hit rates through natively offloadable sparse attention.}

\begin{figure*}[t]
\begin{center}
\includegraphics[width=\linewidth]{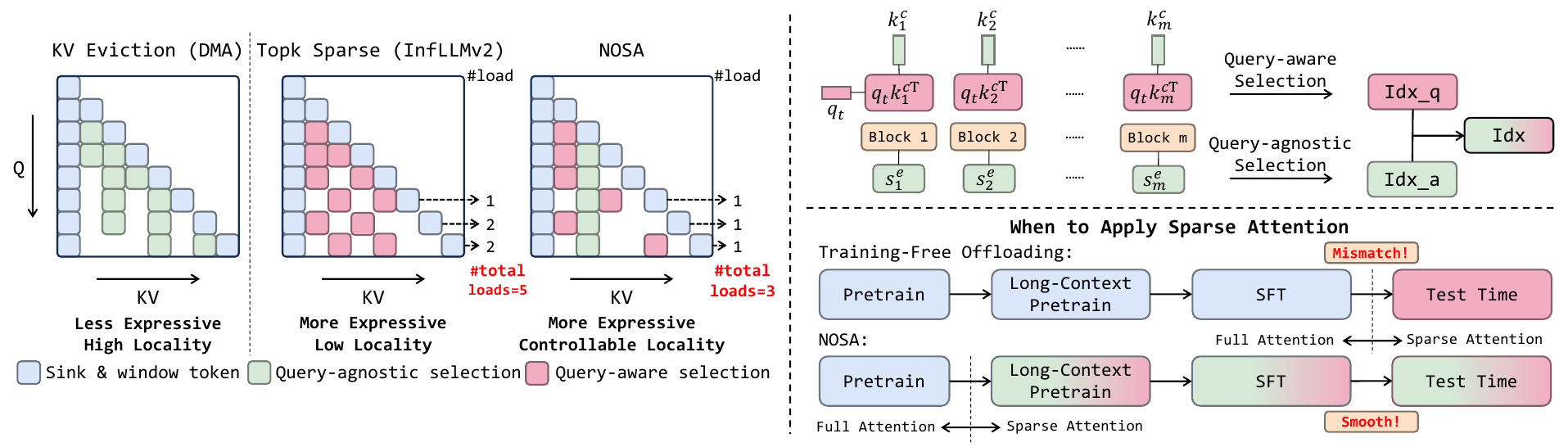}
\end{center}
\vspace{-5pt}
\caption{The framework of \name. 
By introducing a locality constraint, \name~reduces the number of blocks loaded at each step. We apply \name~from the long-context pretraining stage.
}
\vspace{-10pt}
\label{fig:framework}
\end{figure*}

\section{NOSA}

In this section, we first describe the algorithmic design of \name, followed by detailed ablations on core designs. The overall framework of \name~is illustrated in Figure~\ref{fig:framework}.

\subsection{Offloading-aware Trainable Sparse Attention}

\textbf{Adding Locality Constraint.}
As discussed in Section~\ref{sec:observations}, improving locality in token (or block) selection is the key to achieving an efficient offloading system. 
Accordingly, the core idea of \name~is to improve selection locality during both training and inference. A natural approach is to enforce a minimum level of KV selection locality, formalized as follows.
Aiming to maintain a desired locality $\gamma(t)$ across all token positions $t$, we impose a lower bound $\gamma_0 \in (0, 1)$ on the overlapping rate of selected tokens between adjacent decoding steps, i.e., we design an algorithm such that
\begin{align}
    \forall t \in \{2, \cdots, n\}, \gamma(t)\geq \gamma_0.
\end{align}

To achieve this goal, we divide the selected tokens $\Gamma(t)$ into two subsets: the query-aware selection $\Gamma_q(t)$ and the query-agnostic selection $\Gamma_e(t)$. 
For query-aware selection $\Gamma_q(t)$, we do not add any constraints to preserve the model's performance.
As described in Section~\ref{sec:selection}, query-agnostic selection naturally imposes an eviction constraint, i.e., if a token is not selected at decoding step $t$, it must not be selected at any subsequent step $t' > t$, and $\Gamma_e(t')\subseteq\Gamma_e(t)\cup\{t+1, \cdots, t'\}$, which emerges naturally rather than being explicitly designed.

We implement the query-agnostic selection by using an \textit{eviction head} to assign each KV pair an \textit{importance score} $s_t^e \in \mathbb{R}$. When selecting $k$ query-agnostic tokens, a Top-$k$ function is applied to these scores, formulated as $\Gamma_e(t) = \text{ArgTop}_k(\{s_i^e\}_{i=0}^t)$. 
To further enhance performance, we implement the eviction head using DMA~\citep{shi2025trainable}, a trainable sparse attention mechanism with query-agnostic selection, which can be viewed as a trainable KV eviction method. 
The importance score is computed as
\begin{align}
\label{eq:importance_score}
s_j^e = \tau(\mathbf{v}_j \mathbf{W}_1)\mathbf{W}_2,
\end{align}
where $\tau$ is a non-linear function, $\mathbf{W}_1\in \mathbb{R}^{d_{\text{kv}}\times n_{\text{kv}}},~\mathbf{W}_2\in\mathbb{R}^{n_{\text{kv}}\times 1}$ are trainable parameters, and $d_{\text{kv}}, n_{\text{kv}}$ are the dimension and the number of key/value heads, correspondingly. $\tau$ is implemented as the \texttt{softplus} function in DMA's design. 
We remove the final $\exp$ operation from the original design to improve performance, as validated by the ablation studies in Section~\ref{sec:abl_alg}.
For the query-aware selection, we adopt $s_{tj}^q = \mathbf{q}_t \mathbf{k}_j^\top$ as the selection score to preserve the model's ability to retrieve previously evicted tokens, and also apply a Top-$k$ function to these scores for selection. 

\textbf{Adapting to Block-wise Selection.}
Mainstream trainable sparse attention mechanisms can be broadly categorized into element-wise~\citep{deepseekai2024deepseekv32} and block-wise approaches~\citep{yuan2025native, zhao2025infllm}.
The computation of $s_j^e$ and $s_{tj}^q$ described above follows an element-wise paradigm, where selection is performed at the token level.
However, compared to block-wise patterns, element-wise selection is fundamentally inefficient for KV cache offloading due to low PCIe bandwidth utilization, as shown in the ablation studies (Section~\ref{sec:abl_alg}).
Therefore, we adopt a two-stage, block-wise selection strategy and build \name{} on top of \textsc{InfLLMv2}~\citep{zhao2025infllm}. Specifically, mean pooling is first applied to sub-blocks within each block, followed by max pooling at the block level.

Denote the $j$-th block as $[j]$, and each block contains $n_b$ tokens. We first perform mean pooling with a small stride $l_{s}^\text{mean}$ and a kernel size $l_k^\text{mean}$ on each sub-block. We denote the $j$-th sub-block as $\langle j\rangle$, and proceed as follows:
\begin{equation}
    \begin{aligned}\small
    s_{i\langle j\rangle}^q = \frac{1}{l_k^\text{mean}}\sum_{t=j\times l_s^\text{mean}}^{j\times l_s^\text{mean} + l_k^\text{mean}}
    s_{it}^q,~
    s_{\langle j\rangle}^e = \frac{1}{l_k^\text{mean}}\sum_{t=j\times l_s^\text{mean}}^{j\times l_s^\text{mean} + l_k^\text{mean}}s_t^e
    \end{aligned}
\end{equation}
Then, a max pooling is applied to each block:
\begin{equation}
    \begin{aligned}
    s_{i[j]}^q = \max_{\langle t \rangle\subset [j]} s_{i\langle t\rangle}^q,~ 
    s_{[j]}^e = \max_{\langle t\rangle \subset [j]} s_{\langle t\rangle}^e
    \end{aligned}
\end{equation}
At each decoding step, $k/n_b$ blocks are selected. We assign two budgets $k_q, k_e$ for query-aware and query-agnostic selection, such that $k=k_q+k_e$. We first select the query-aware blocks by $s_{i[j]}^q$. We then set these places to infinity and select the query-agnostic blocks by $s_{[j]}^e$. Such process follows taking a Top-$k$ selection ($\Gamma(i) = \text{ArgTop}_{k/n_b}(\{s_{i[j]}\}_{j=1}^{\lfloor t/n_b\rfloor})$) on the following score:
\begin{align}
\label{eq:selection}
    s_{i[j]} =
    \begin{cases}
        +\infty, & s_{[j]}^q\in \text{Top}_{k_q/n_b}\left(\{s_{[j]}^q\}_{j=1}^{\lfloor i/n_b\rfloor}\right);\\
        s_{[j]}^e, & \text{otherwise.}
    \end{cases}
\end{align}

This design guarantees a lower bound on locality, as formalized in Theorem~\ref{theorem:locality} (Proved in Appendix~\ref{sec:locality_theorem}).
\begin{theorem}
\label{theorem:locality}
If the above selection process has budget $k=k_q+k_e$, we have $\forall t\in \{2, \cdots, n\},~\gamma(t) \geq \frac{k_e}{k}$.
\end{theorem}

\textbf{Attention Calculation.}
At the decoding step for generating the $(t+1)$-th token, the hidden input is $\textbf{h}_t \in \mathbb{R}^{1 \times d}$, with its corresponding query, key, and value vectors $\textbf{q}_t, \textbf{k}_t, \textbf{v}_t \in \mathbb{R}^{1 \times d}$, and the KV cache $\mathbf{K}, \mathbf{V} \in \mathbb{R}^{t \times d}$. We first compute the importance scores $\mathbf{S}^{e} = (s_j^e)_{t}$ according to Equation~\ref{eq:importance_score}. 
Next, we utilize the selection process defined in Equation~\ref{eq:selection} to generate the selected positions $\Gamma(t)$ and the corresponding attention mask $\mathbf{M}$, where $m_{j} = 0$ if $j \in \Gamma(t)$ and $m_{j} = -\infty$ otherwise. To make this selection differentiable and trainable, we introduce an attention bias vector $\mathbf{b} \in \mathbb{R}^t$ similar to~\citet{shi2025trainable}, where each element is assigned as $b_j = {s}_{j}^e = \tau(\mathbf{v}_j \mathbf{W}_1)\mathbf{W}_2$.

Finally, the attention output $\mathbf{o}_t \in \mathbb{R}^d$ is computed as
\begin{align}
    \textbf{o}_t = \sum_{j=1}^{t} \frac{\exp(b_j) \exp(\mathbf{q}_t\mathbf{k}_j^\top + m_j)}{\sum_{i=1}^t\left(\exp(b_i) \exp(\mathbf{q}_t\mathbf{k}_i^\top + m_i)\right)}\mathbf{v}_j.
\end{align}

We observe that the query-agnostic selection is highly sensitive to numerical precision. Therefore, we omit the exponential ($\exp$) operation before token selection; that is, we select the top-$k$ indices based on pre-exponential importance score ${s}_{[j]}^e$ rather than on $\exp(b_j)$. Since this operation is deferred to the attention computation, we refer to this optimization as ED-DMA (Exp-Delayed DMA). As shown in the ablation studies in Section~\ref{sec:abl_alg}, ED-DMA yields the most stable and effective results, whereas alternative approaches lead to noticeably greater performance degradation.

\textbf{Training with \name.}
The design of \name~is fully trainable, as the eviction head receives gradients through the attention bias $\mathbf{b}$. To equip LLMs with native offloading capability, \textit{we incorporate \name~throughout long-context adaptation}. Specifically, we pretrain with vanilla attention on short-context data, then switch to \name~for long-context continual pretraining. We also apply \name~during SFT, therefore delivering the model with built-in \name~support.

\begin{wrapfigure}{r}{0.55\textwidth}
\vspace{-10pt} 
\centering
\small

\scalebox{0.74}{
\begin{tabular}{l|c|ccc>{\columncolor{pink!20}}c}
\toprule
Implementation & \textsc{InfLLMv2} & \textsc{Locret} & \textsc{DMA} & \textsc{S-DMA} & \textbf{\textsc{ED-DMA}} \\
\midrule
RULER-16K (\%) $\uparrow$ & 60.3 & 55.9 & 56.2 & 60.9 & \textbf{61.9}\\
\bottomrule
\end{tabular}
}
\captionof{table}{Various eviction head implementations.}
\label{tab:results-abl-eviction-head}

\vspace{6pt}

\begin{subfigure}{0.48\linewidth}
    \centering
    \includegraphics[width=\linewidth]{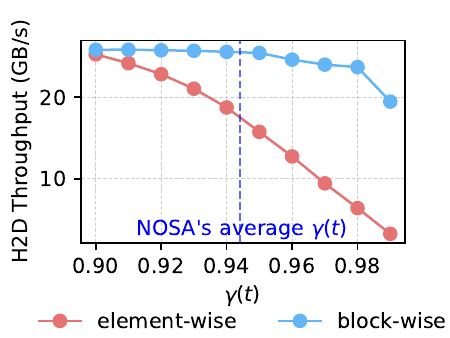}
    \caption{Block-wise vs. element-wise host-to-device throughput}
    \label{fig:element-comm}
\end{subfigure}
\hfill
\begin{subfigure}{0.48\linewidth}
    \centering
    \includegraphics[width=\linewidth]{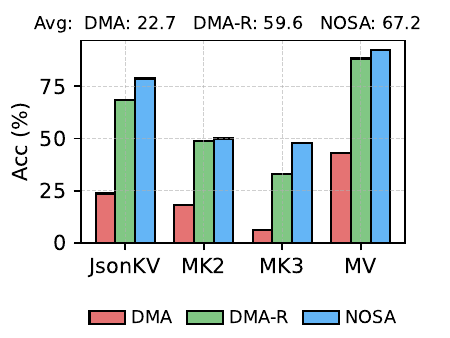}
    \caption{DMA, DMA-R, and \name's accuracies on HELMET's recall.}
    \label{fig:abl-helmet-recall}
\end{subfigure}

\captionof{figure}{Comparison between DMA and NOSA.}
\label{fig:dma-and-nosa}

\vspace{-10pt} 
\end{wrapfigure}

\subsection{Ablations on Algorithm Design}
\label{sec:abl_alg}

We ablate the core components of \name~in this section.

\textbf{Eviction Head Implementation.}
To demonstrate that ED-DMA is the most suitable implementation for the eviction head, we compare ED-DMA with several alternative designs, including \textsc{Locret}~\citep{huang2024locret}, vanilla DMA~\citep{shi2025trainable}, and simple-DMA. 
Specifically, \textsc{Locret} employs a lightweight trainable MLP to produce the importance score, i.e., $s_j^e = \text{MLP}(\mathbf{h}_j)$. 
Vanilla DMA selects important blocks based on $\exp(b_j)$ without delaying the $\exp$ operation. 
Simple-DMA (S-DMA) applies a straight-through estimator to the attention bias $b_j$ in ED-DMA; as a result, its attention computation is identical to vanilla attention.
More details on each implementation and the full results are included in Appendix~\ref{sec:abl-eviction-head}.

We apply these implementations and conduct long-context continual pretraining on a 1B-size model, then evaluate on RULER~\citep{hsieh2024ruler} with 16K input lengths.
As shown in Table~\ref{tab:results-abl-eviction-head}, ED-DMA outperforms all other implementations. Notably, the large performance degradation of vanilla DMA indicates the significant negative impact of the precision loss introduced by the $\exp$ operation.

\textbf{Locality.}
We compare the locality $\gamma(t)$ for each transformer layer of the trained 1B-size model using an input sequence of 16K tokens sampled from PG19~\citep{raecompressive2019}. As shown in Figure~\ref{fig:infllmv2_locality}, \name~obtains a higher locality compared with the original \textsc{InfLLMv2}, yielding an approximate $5.5\%$ improvement (equivalent to nearly a $2\times$ reduction in cache miss rate in KV cache offloading systems). These results demonstrate the effectiveness of incorporating a locality constraint to enhance locality.

\textbf{Comparison with DMA.} 
Since the eviction head in \name~is inspired by DMA, we provide an intuitive discussion of how our design differs from DMA.
Specifically, (1) DMA performs element-wise selection, whereas \name~selects KV cache in a block-wise manner; and (2) \name~uses query-aware selection with query-agnostic selection, while DMA only relies on query-agnostic selection.

Element-wise selection is ill-suited for KV cache offloading. The large number of irregular offsets and highly fragmented host-to-device (CPU to GPU) transfers result in poor PCIe bandwidth utilization. 
We compare the two communication patterns under batch size $B=64$, sequence length $n=4096$, head dimension $d_{\text{kv}}=128$, and $n_{\text{kv}}=2$ KV heads, across different locality levels $\gamma(t)$. We only fetch KV units that are not resident on the GPU, i.e., a fraction of $1-\gamma(t)$. Both kernels are implemented in \texttt{Triton} and use UVA to directly access host memory. 
Figure~\ref{fig:element-comm} shows that element-wise communication degrades sharply under high locality, whereas block-wise communication sustains high throughput, motivating our use of the block-wise pattern.

We introduce query-aware selection in \name~because DMA's query-agnostic policy is inadequate for long-context recall, as it causes irreversible eviction. 
We evaluate DMA and \name~on HELMET recall tasks~\citep{yen2024helmet} (MK2, MK3, MV from RULER~\citep{hsieh2024ruler}, and JsonKV), and also include DMA-R (trained with DMA, inference with \name).
Figure~\ref{fig:abl-helmet-recall} shows DMA has low recall. 
Query-aware selection at test time improves significantly, and \name~further outperforms DMA-R, suggesting query-aware selection is necessary in both training and inference.

\begin{figure}[t]
    \centering
    \begin{minipage}{0.49\linewidth}
        \centering
        \includegraphics[width=\linewidth]{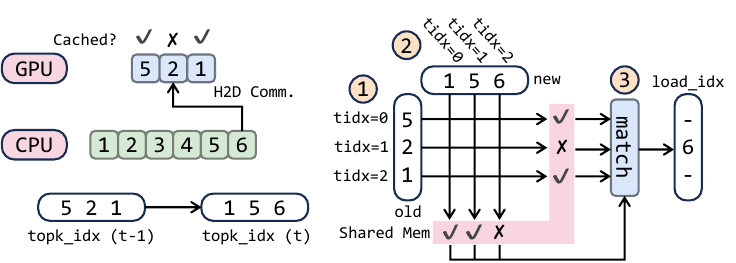}
        \vspace{-5pt}
        \captionof{figure}{Memory layout and block selection.}
        \vspace{-15pt}
        \label{fig:block_selection}
    \end{minipage}\hfill
    \begin{minipage}{0.24\linewidth}
        \centering
        \includegraphics[width=0.95\linewidth]{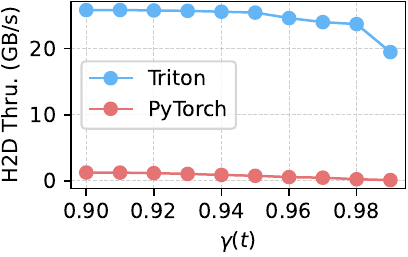}
        \vspace{-5pt}
        \captionof{figure}{Throughput ($\uparrow$) of H2D communication.}
        \vspace{-15pt}
        \label{fig:h2d_thru}
    \end{minipage}\hfill
    \begin{minipage}{0.24\linewidth}
        \centering
        \includegraphics[width=0.95\linewidth]{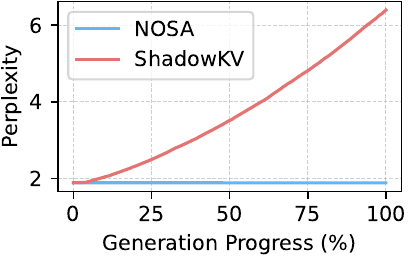}
        \vspace{-5pt}
        \captionof{figure}{Perplexity ($\downarrow$) in long-generation.}
        \vspace{-15pt}
        \label{fig:ppl}
    \end{minipage}
\end{figure}


\section{NOSI: Inference System for NOSA}

The vanilla implementation of \name~based on Hugging Face Transformers~\citep{wolf-etal-2020-transformers} is highly inefficient due to two factors due to inefficient kernels and excessive small kernel launch overheads.
To mitigate these issues, we implement \namei, a \underline{NO}SA-based \underline{S}ystem for \underline{I}nference, which exposes true communication overhead from other performance bottlenecks. Below, we briefly introduce our optimizations. We provide a pseudocode of \namei's implementation of \name~inference in Appendix~\ref{sec:pseudocode}.

\textbf{Kernel Fusion.}
We utilize \textsc{FlashInfer}'s~\citep{ye2025flashinfer} implementations for LayerNorm, RoPE, and FFN. Since the eviction heads have fewer parameters, running them alone incurs substantial overhead; therefore, we fuse the eviction heads with splitting QKV tensor into a single kernel. As the max-pooling stage is applied identically to both query-aware and query-agnostic scores, we also fuse the two poolings into one kernel. Both fused kernels are hand-crafted in \texttt{Triton}. We further use \texttt{CUDA Graphs} to reduce the kernel launch overhead incurred by the process from max-pooling to generating the selected indexes.

\textbf{Memory Layout and Block Selection.}
We organize the KV cache into minimal block-sized units on both GPU and CPU (Figure~\ref{fig:block_selection}) to reduce communication overhead. Since decoding does not require KV blocks on GPU to be contiguous, we only swap out blocks not used in the current step and replace them with the newly selected ones. This requires solving a set-difference problem and producing a replacement mapping.
Since implementing this efficiently in \texttt{Pytorch} is hard, we use a custom \texttt{CUDA} kernel (Figure~\ref{fig:block_selection}). For each thread, the kernel runs in $O(k)$, as it checks whether each old index appears in the new set, then uses shared memory to test whether new indices already exist in the old set, and finally performs a per-thread sequential pass to generate the final mapping, which is parallel-friendly.

\textbf{Offloading Communication.} 
Since \name~selects blocks independently for each attention head, each block is relatively small (16 KiB in our setting). This fine-grained partitioning results in fragmented host-to-device communication and poor PCIe bandwidth utilization. To mitigate this issue, inspired by~\citet{zhou2025sparseserve}, we implement a custom \texttt{Triton} kernel for host-to-device transfers and directly leverage Unified Virtual Addressing (UVA) to access CPU memory.

We compare our kernel implementation with the vanilla \texttt{PyTorch} implementation. As shown in Figure~\ref{fig:h2d_thru}, our method consistently achieves substantially higher throughput than the \texttt{PyTorch} baseline. Furthermore, we evaluate the communication throughput in real-world inference scenarios using an 8B model with batch sizes $B\in\{16,32,64\}$ on a PCIe 4.0 platform. The real-world communication throughput under these settings are 17.04, 20.36, and 22.35 GB/s, demonstrating the efficiency and scalability of our implementation.

\begin{table*}[t]
\small

\centering
\setlength{\tabcolsep}{0.7pt} 
\scalebox{0.74}{
\begin{tabular}{l!{\vrule width \heavyrulewidth}ccccccccccccc|c!{\vrule width \heavyrulewidth}ccccccc|c}
\toprule
\multirow{2}{*}{Method} & \multicolumn{14}{c!{\vrule width \heavyrulewidth}}{\textbf{\textit{LongBench}}} & \multicolumn{8}{c}{\textbf{\textit{Helmet}}}\\
& GO & TQ & NQ & QS & MU & 2W & MQ & RB & HQ & TR & PR & PC & SA & Avg. $\uparrow$ & Recall & RAG & ICL & Cite & Rerank & LongQA & Summ. & Avg. $\uparrow$\\
\midrule
\multicolumn{23}{c}{\small\textit{1B Model\quad Full Prefill}} \\
\midrule
\textsc{FullAttn} & 28.0 & 70.8 & 16.0 & 23.7 & 11.5 & 19.5 & 45.2 & 55.0 & 21.0 & 66.5 & 10.5 & 1.0 & 23.55 & 30.2 & 61.3$_{\pm0.4}$ & 39.0$_{\pm0.9}$ & 60.0$_{\pm0.7}$ & 5.5$_{\pm0.3}$ & 14.3$_{\pm0.0}$ & 17.2$_{\pm0.3}$ & 0.4$_{\pm0.2}$ & 28.3$_{\pm0.3}$
\\
\midrule
\textsc{ShadKV}$_{64}$ & 25.2 & 70.1 & 14.9 & 23.0 & 10.7 & 18.8 & \textbf{43.4} & 54.6 & 19.2 & 67.0 & 12.0 & 0.0 & \textbf{22.6} & 29.4 & 25.9$_{\pm1.0}$ & 37.5$_{\pm1.3}$ & 55.4$_{\pm0.5}$ & 2.7$_{\pm0.4}$ & 5.8$_{\pm0.0}$ & 16.4$_{\pm0.3}$ & 0.1$_{\pm0.1}$ & 20.6$_{\pm0.2}$ \\
\textsc{ShadKV}$_{8}$ & 25.0 & 70.3 & \textbf{15.4} & \textbf{23.1} & \textbf{11.1} & \textbf{19.7} & 42.3 & 55.2 & \textbf{19.6} & 66.5 & 11.0 & \textbf{1.0} & \textbf{22.6} & 29.4 & \textbf{48.9}$_{\pm0.8}$ & 38.4$_{\pm1.3}$ & 59.0$_{\pm0.4}$ & \textbf{4.1}$_{\pm0.3}$ & 11.0$_{\pm0.0}$ & 16.7$_{\pm0.3}$ & 0.1$_{\pm0.1}$ & \textbf{25.5}$_{\pm0.3}$ \\
\textsc{ArkVale} & 8.2 & 58.7 & 8.4 & 11.4 & 9.5 & 13.8 & 25.2 & 38.9 & 10.5 & 60.5 & 9.8 & \textbf{1.0} & 11.6 & 20.6 & 26.7$_{\pm1.6}$ & 34.3$_{\pm1.0}$ & 57.5$_{\pm0.3}$ & 1.1$_{\pm1.0}$ & 1.8$_{\pm0.4}$ & 16.9$_{\pm0.7}$ & 0.2$_{\pm0.1}$ & 19.8$_{\pm0.3}$ \\
\textsc{DMA}$^F$ & 23.8 & 79.2 & 13.5 & 22.4 & 7.9 & 15.3 & 38.0 & 54.4 & 17.3 & \textbf{68.5} & 10.0 & 0.5 & 21.3 & 28.6 & 15.9$_{\pm1.0}$ & 39.6$_{\pm1.1}$ & 54.7$_{\pm0.4}$ & 2.7$_{\pm0.1}$ & 8.7$_{\pm0.0}$ & 15.8$_{\pm0.7}$ & \textbf{0.8}$_{\pm0.3}$ & 19.7$_{\pm0.3}$\\
\rowcolor{pink!20}
\textbf{\name$^F$} & \textbf{28.8} & \textbf{79.8} & 12.9 & 22.9 & 8.3 & 14.5 & 41.8 & \textbf{55.6} & 18.7 & 68.0 & \textbf{14.5} & \textbf{1.0} & 20.3 & \textbf{29.8} & 34.3$_{\pm1.0}$ & \textbf{39.8}$_{\pm1.1}$ & \textbf{59.2}$_{\pm0.6}$ & 3.5$_{\pm0.1}$ & \textbf{12.7}$_{\pm0.0}$ & \textbf{17.6}$_{\pm1.0}$ & \textbf{0.8}$_{\pm0.2}$ & 24.0$_{\pm0.3}$\\
\midrule
\multicolumn{23}{c}{\small\textit{1B Model\quad Sparse Prefill}} \\
\midrule
\textsc{InfLLMv2} & 28.7 & 73.3 & 15.0 & 23.1 & 8.7 & 19.8 & 42.7 & 55.5 & 18.6 & 67.5 & 9.5 & 2.0 & 23.0 & 29.8 & 37.2$_{\pm1.4}$ & 37.2$_{\pm0.6}$ & 58.0$_{\pm0.1}$ & 4.1$_{\pm0.6}$ & 15.5$_{\pm0.0}$ & 17.7$_{\pm0.3}$ & 0.8$_{\pm0.3}$ & 24.4$_{\pm0.1}$\\
\midrule
\textsc{ShadKV}$_{64}^M$ & 26.0 & 73.2 & 15.0 & \textbf{23.8} & 8.1 & \textbf{18.9} & 42.0 & 54.0 & \textbf{19.3} & 66.5 & 10.5 & 1.5 & 20.5 & 29.2 & 26.4$_{\pm0.9}$ & 37.9$_{\pm1.2}$ & 50.1$_{\pm0.3}$ & 3.4$_{\pm0.0}$ & 9.6$_{\pm0.0}$ & 17.3$_{\pm0.5}$ & 0.2$_{\pm0.1}$ & 20.7$_{\pm0.2}$\\
\textsc{ShadKV}$_{8}^M$ & 25.3 & 73.1 & 15.7 & 23.6 & 9.1 & 18.6 & 42.1 & 54.0 & 19.1 & 66.5 & 9.5 & \textbf{2.0} & \textbf{21.9} & 29.3 & 
\textbf{41.9}$_{\pm0.6}$ & {38.5}$_{\pm0.9}$ & 53.9$_{\pm0.4}$ & \textbf{4.2}$_{\pm0.2}$ & 11.2$_{\pm0.0}$ & \textbf{17.5}$_{\pm0.5}$ & 0.1$_{\pm0.1}$ & \textbf{23.9}$_{\pm0.2}$\\
\textsc{InfLLM}$_{128}$ & 27.9 & 67.3 & 14.7 & 23.2 & 9.2 & 17.5 & 42.0 & 52.6 & {19.0} & 63.5 & 5.0 & 1.5 & 19.4 & 27.9 & 4.4$_{\pm0.3}$ & 34.4$_{\pm0.3}$ & 48.8$_{\pm0.5}$ & 1.2$_{\pm0.3}$ & 2.6$_{\pm0.0}$ & 13.2$_{\pm0.3}$ & 0.2$_{\pm0.1}$ & 15.0$_{\pm0.1}$\\
\textsc{InfLLM}$_{64}$ & 28.0 & 68.2 & \textbf{15.3} & 22.7 & 8.8 & 17.9 & \textbf{43.0} & 52.4 & 17.8 & 64.5 & 5.5 & 1.0 & 18.6 & 28.0 & 3.7$_{\pm0.3}$ & 36.4$_{\pm0.4}$ & 51.7$_{\pm0.3}$ & 1.1$_{\pm0.2}$ & 2.0$_{\pm0.0}$ & 13.0$_{\pm0.6}$ & 0.4$_{\pm0.2}$ & 15.5$_{\pm0.2}$ \\
\textsc{DMA}$^S$ & 23.8 & 77.2 & 13.1 & 22.1 & 7.3 & 15.7 & 37.9 & 52.8 & 17.1 & 67.0 & 10.5 & 0.5 & 19.4 & 28.0 & 14.5$_{\pm0.8}$ & 28.2$_{\pm0.7}$ & 48.7$_{\pm0.3}$ & 2.8$_{\pm0.3}$ & 7.6$_{\pm0.0}$ & 14.6$_{\pm0.7}$ & 0.4$_{\pm0.1}$ & 16.7$_{\pm0.1}$\\
\rowcolor{pink!20}
\textbf{\name$^S$} & \textbf{28.3} & \textbf{81.0} & 14.2 & 22.7 & \textbf{9.9} & 15.3 & 42.9 & \textbf{54.6} & 17.9 & \textbf{68.5} & \textbf{16.5} & 1.5 & 20.2 & \textbf{30.3} & 29.2$_{\pm1.0}$ & \textbf{39.6}$_{\pm0.3}$ & \textbf{57.5}$_{\pm0.4}$ & 3.6$_{\pm0.4}$ & \textbf{13.6}$_{\pm0.0}$ & 16.2$_{\pm0.7}$ & \textbf{0.6}$_{\pm0.1}$ & 22.9$_{\pm0.2}$\\
\specialrule{\heavyrulewidth}{0.5ex}{0.5ex}
\multicolumn{23}{c}{\small\textit{3B Model\quad Full Prefill}} \\
\midrule
\textsc{FullAttn} & 31.1 & 85.8 & 20.5 & 23.8 & 15.7 & 27.0 & 50.4 & 49.4 & 31.0 & 74.0 & 76.5 & 2.5 & 6.9 & 38.0 & 48.4$_{\pm1.1}$ & 50.6$_{\pm0.2}$ & 49.5$_{\pm0.5}$ & 9.0$_{\pm0.2}$ & 23.1$_{\pm0.0}$ & 21.5$_{\pm0.6}$ & 5.0$_{\pm0.2}$ & 29.6$_{\pm0.3}$\\
\midrule
\textsc{ShadKV}$_{64}$ & 27.5 & 84.2 & 20.3 & 23.3 & 14.6 & 24.3 & 49.3 & 48.4 & 31.5 & 73.5 & 64.0 & \textbf{2.5} & 6.4 & 36.1 & 23.1$_{\pm1.2}$ & 48.3$_{\pm0.6}$ & 46.4$_{\pm0.6}$ & 4.4$_{\pm0.4}$ & 11.6$_{\pm0.0}$ & 20.6$_{\pm0.4}$ & 6.2$_{\pm0.4}$ & 22.9$_{\pm0.3}$\\
\textsc{ShadKV}$_{8}$ & 27.4 & 83.8 & 20.1 & 23.3 & 14.5 & 24.2 & 48.6 & 49.7 & 31.1 & 73.5 & 75.0 & \textbf{2.5} & 5.9 & 36.9 & 39.4$_{\pm0.7}$ & 49.0$_{\pm0.3}$ & 48.0$_{\pm0.9}$ & 5.5$_{\pm0.4}$ & 15.0$_{\pm0.0}$ & 21.2$_{\pm0.8}$ & \textbf{7.3}$_{\pm0.2}$ & 26.5$_{\pm0.3}$\\
\textsc{ArkVale} & 30.3 & 86.1 & \textbf{20.5} & 23.1 & 15.1 & 27.7 & 50.0 & 48.1 & 31.7 & 74.0 & \textbf{77.0} & \textbf{2.5} & 6.9 & \textbf{37.9} & 28.4$_{\pm0.9}$ & \textbf{50.4}$_{\pm0.3}$ & \textbf{49.3}$_{\pm0.6}$ & 5.9$_{\pm0.5}$ & 21.1$_{\pm0.6}$ & 22.1$_{\pm0.1}$ & 5.7$_{\pm0.6}$ & 26.1$_{\pm0.2}$\\
\textsc{DMA}$^F$ & 28.1 & 85.6 & 18.1 & 21.7 & 9.1 & 29.1 & 40.4 & 48.1 & 30.1 & 69.5 & 22.0 & 0.5 & 4.1 & 33.5 & 12.6$_{\pm0.1}$ & 47.2$_{\pm0.7}$ & 47.6$_{\pm1.2}$ & 2.8$_{\pm0.3}$ & 16.0$_{\pm0.0}$ & \textbf{23.9}$_{\pm0.7}$ & 4.4$_{\pm0.1}$ & 22.1$_{\pm0.3}$\\
\rowcolor{pink!20}
\textbf{\name$^F$} & \textbf{31.7} & \textbf{87.2} & 20.2 & \textbf{23.7} & \textbf{18.1} & \textbf{32.7} & \textbf{52.1} & \textbf{50.2} & \textbf{41.2} & \textbf{75.5} & 45.0 & 1.0 & \textbf{7.2} & 37.4 & \textbf{45.1}$_{\pm0.3}$ & 49.8$_{\pm0.3}$ & 47.7$_{\pm0.3}$ & \textbf{6.3}$_{\pm0.4}$ & \textbf{24.3}$_{\pm0.0}$ & 21.5$_{\pm0.8}$ & 2.7$_{\pm0.3}$ & \textbf{28.2}$_{\pm0.1}$\\
\midrule
\multicolumn{23}{c}{\small\textit{3B Model\quad Sparse Prefill}} \\
\midrule
\textsc{InfLLMv2} & 31.3 & 87.2 & 19.7 & 23.2 & 14.7 & 23.8 & 49.4 & 48.4 & 30.1 & 73.5 & 83.0 & 3.0 & 7.6 & 38.1 & 39.7$_{\pm0.6}$ & 49.4$_{\pm0.4}$ & 48.6$_{\pm0.9}$ & 5.7$_{\pm1.3}$ & 22.7$_{\pm0.0}$ & 23.0$_{\pm1.1}$ & 3.2$_{\pm0.5}$ & 27.5$_{\pm0.5}$\\
\midrule
\textsc{ShadKV}$_{64}^M$ & 28.2 & 84.6 & 19.7 & 23.1 & 14.2 & 25.7 & 48.1 & 47.2 & 30.5 & 72.0 & 62.5 & \textbf{3.0} & 5.9 & 35.7 & 27.0$_{\pm0.6}$ & 48.0$_{\pm1.1}$ & 39.1$_{\pm0.7}$ & 5.0$_{\pm0.9}$ & 13.5$_{\pm0.2}$ & {21.2}$_{\pm1.1}$ & \textbf{5.6}$_{\pm0.4}$ & 22.8$_{\pm0.6}$\\
\textsc{ShadKV}$_{8}^M$ & 27.9 & 84.5 & 19.4 & 23.1 & 14.1 & 25.4 & 48.0 & 48.5 & 29.9 & 72.5 & \textbf{67.0} & \textbf{3.0} & 6.2 & 36.1 & 
\textbf{43.7}$_{\pm0.5}$ & 48.9$_{\pm1.0}$ & 40.5$_{\pm0.6}$ & 7.4$_{\pm0.7}$ & 17.1$_{\pm0.0}$ & 21.1$_{\pm1.0}$ & \textbf{5.6}$_{\pm0.7}$ & {26.3}$_{\pm0.3}$\\
\textsc{InfLLM}$_{128}$ & 30.1 & 83.5 & 18.5 & 22.3 & 14.0 & 27.3 & 48.5 & 47.5 & 33.2 & 71.0 & 37.0 & 2.5 & 6.4 & 34.0 & 5.5$_{\pm1.4}$ & 44.2$_{\pm0.4}$ & 35.6$_{\pm0.5}$ & 1.7$_{\pm0.4}$ & 2.8$_{\pm0.0}$ & 12.7$_{\pm1.8}$ & \textbf{5.6}$_{\pm0.4}$ & 15.4$_{\pm0.3}$\\
\textsc{InfLLM}$_{64}$ & 29.9 & 82.8 & 17.9 & 22.3 & 15.6 & 26.6 & 47.9 & 46.9 & 31.9 & 73.0 & 41.5 & 1.5 & 5.8 & 34.1 & 4.5$_{\pm0.4}$ & 44.1$_{\pm0.5}$ & 37.5$_{\pm0.2}$ & 1.5$_{\pm0.5}$ & 1.9$_{\pm0.0}$ & 12.0$_{\pm1.9}$ & 4.9$_{\pm0.6}$ & 15.2$_{\pm0.2}$\\
\textsc{DMA}$^S$ & 28.1 & 85.6 & 18.1 & 21.7 & 9.1 & 29.1 & 40.4 & 48.1 & 30.1 & 69.5 & 22.0 & 0.5 & 4.1 & 31.3 & 12.3$_{\pm0.6}$ & 41.2$_{\pm0.9}$ & 41.1$_{\pm0.0}$ & 3.6$_{\pm0.4}$ & 10.1$_{\pm0.0}$ & 21.5$_{\pm0.8}$ & 5.3$_{\pm0.6}$ & 19.3$_{\pm0.2}$\\
\rowcolor{pink!20}
\textbf{\name$^S$} & \textbf{31.7} & \textbf{88.2} & \textbf{21.2} & \textbf{23.5} & \textbf{19.7} & \textbf{32.9} & \textbf{52.5} & \textbf{49.4} & \textbf{40.0} & \textbf{74.5} & 52.5 & 0.6 & \textbf{7.2} & \textbf{38.0} & 36.1$_{\pm1.0}$ & \textbf{49.2}$_{\pm0.6}$ & \textbf{52.2}$_{\pm0.4}$ & \textbf{6.4}$_{\pm0.2}$ & \textbf{26.4}$_{\pm0.0}$ & \textbf{21.6}$_{\pm0.7}$ & 2.6$_{\pm1.1}$ & \textbf{27.8}$_{\pm0.3}$\\
\specialrule{\heavyrulewidth}{0.5ex}{0.5ex}
\multicolumn{23}{c}{\small\textit{8B Model\quad Full Prefill}} \\
\midrule
\textsc{FullAttn} & 31.0 & 88.1 & 19.4 & 25.0 & 27.4 & 40.6 & 52.3 & 58.3 & 50.0 & 74.5 & 94.0 & 3.0 & 6.6 & 43.9 &
89.5$_{\pm0.3}$ & 56.1$_{\pm1.0}$ & 61.3$_{\pm0.4}$ & 14.5$_{\pm0.4}$ & 39.8$_{\pm0.0}$ & 26.6$_{\pm0.6}$ & 10.7$_{\pm1.0}$ & 42.6$_{\pm0.3}$
\\
\midrule
\textsc{ShadKV}$_{64}$ & 28.5 & 87.8 & \textbf{19.4} & 24.2 & 25.0 & 41.1 & 52.0 & 57.4 & 48.7 & \textbf{74.5} & 89.5 & 3.0 & \textbf{6.2} & 42.9 
& 55.3$_{\pm0.2}$ & 55.6$_{\pm0.9}$ & 58.9$_{\pm0.6}$ & 10.7$_{\pm0.6}$ &
21.3$_{\pm0.1}$ & 25.2$_{\pm1.2}$ & 7.8$_{\pm0.5}$ & 33.5$_{\pm0.3}$\\
\textsc{ShadKV}$_{8}$ & 28.8 & 87.7 & \textbf{19.4} & 24.4 & 26.2 & \textbf{41.4} & 51.4 & 57.5 & 48.8 & \textbf{74.5} & \textbf{93.0} & 3.0 & 6.1 & 43.3 
& 78.3$_{\pm0.4}$ & 55.7$_{\pm1.1}$ & 60.6$_{\pm0.5}$ & 12.4$_{\pm0.9}$ &
29.3$_{\pm0.0}$ & 26.6$_{\pm0.8}$ & \textbf{8.8}$_{\pm1.3}$ & 38.8$_{\pm0.6}$\\
\textsc{ArkVale} &
29.7 & \textbf{88.4} & 17.6 & 24.6 & 25.0 & 33.7 & 36.2 & 45.9 & 42.7 & 71.0 & 92.5 & 3.0 & 3.4 & 39.5 
& 56.2$_{\pm1.5}$ & 55.6$_{\pm0.8}$ & {61.5}$_{\pm0.6}$ & 10.7$_{\pm2.4}$ &
\textbf{35.1}$_{\pm0.1}$ & 26.5$_{\pm1.5}$ & 8.2$_{\pm0.8}$ & 36.2$_{\pm0.4}$\\
\textsc{DMA}$^F$ & 27.6 & 86.7 & 16.7 & 23.5 & 24.4 & 39.0 & 47.1 & 55.1 & 47.8 & 73.0 & 69.0 & \textbf{6.0} & 5.7 & 40.1 
& 33.1$_{\pm0.2}$ & 52.4$_{\pm0.6}$ & \textbf{63.8}$_{\pm0.1}$ & 10.4$_{\pm0.7}$ & 26.5$_{\pm0.0}$ & 25.7$_{\pm1.9}$ & 6.5$_{\pm0.7}$ & 31.2$_{\pm0.5}$\\
\rowcolor{pink!20}
\textbf{\name$^F$} &
\textbf{31.8} & 86.8 & 15.2 & \textbf{25.1} & \textbf{29.2} & 41.1 & \textbf{53.0} & \textbf{59.7} & \textbf{50.7} & 74.0 & 92.0 & {4.0} & 5.4 & \textbf{43.7} 
& \textbf{86.3}$_{\pm0.0}$ & \textbf{56.3}$_{\pm0.4}$ & 60.6$_{\pm0.5}$ & \textbf{13.6}$_{\pm1.1}$ &
30.8$_{\pm0.0}$ & \textbf{26.7}$_{\pm1.5}$ & 8.5$_{\pm0.7}$ & \textbf{40.4}$_{\pm0.5}$ \\
\midrule
\multicolumn{23}{c}{\small\textit{8B Model\quad Sparse Prefill}} \\
\midrule
\textsc{InfLLMv2} & 32.4 & 86.6 & 17.7 & 25.0 & 29.3 & 42.2 & 52.9 & 59.5 & 48.7 & 75.5 & 95.5 & 4.5 & 5.8 & 44.3 
& 72.7$_{\pm0.2}$ & 54.1$_{\pm1.1}$ & 68.9$_{\pm0.3}$ & 13.8$_{\pm0.7}$ &
37.5$_{\pm0.0}$ & 26.6$_{\pm1.5}$ & 8.6$_{\pm0.6}$ & 40.3$_{\pm0.3}$\\
\midrule
\textsc{ShadKV}$_{64}^M$ & 28.3 & 89.5 & 18.1 & 24.0 & 23.6 & 37.8 & \textbf{52.9} & 56.6 & \textbf{49.0} & 72.5 & 74.5 & 3.0 & 4.9 & 41.1
& 54.5$_{\pm0.9}$ & 53.1$_{\pm0.7}$ & 52.4$_{\pm0.5}$ & 10.7$_{\pm1.1}$ &
23.4$_{\pm0.0}$ & 23.9$_{\pm1.1}$ & 8.1$_{\pm0.6}$ & 32.3$_{\pm0.4}$\\
\textsc{ShadKV}$_{8}^M$ & 28.3 & \textbf{89.6} & 17.4 & 23.8 & 24.3 & 37.6 & 52.4 & 56.7 & 48.6 & 72.5 & 75.5 & 2.5 & 4.8 & 41.1
& \textbf{74.6}$_{\pm0.2}$ & 53.1$_{\pm0.7}$ & 53.5$_{\pm0.2}$ & 11.6$_{\pm0.5}$ &
28.9$_{\pm0.0}$ & {25.9}$_{\pm2.1}$ & 7.7$_{\pm1.0}$ & 36.5$_{\pm0.4}$\\
\textsc{InfLLM}$_{128}$ & 30.4 & 87.0 & 18.7 & 22.8 & 31.8 & 40.7 & 50.1 & \textbf{60.6} & 48.2 & 71.5 & 49.0 & \textbf{7.5} & 3.4 & 40.1
& 5.8$_{\pm0.1}$ & 47.2$_{\pm0.5}$ & 36.7$_{\pm0.9}$ & 1.7$_{\pm0.1}$ &
2.2$_{\pm0.0}$ & 16.6$_{\pm2.5}$ & 8.1$_{\pm0.4}$ & 16.9$_{\pm0.3}$\\
\textsc{InfLLM}$_{64}$ & 30.2 & 86.2 & \textbf{19.0} & 23.1 & \textbf{33.8} & 38.8 & 50.2 & 60.4 & \textbf{49.0} & \textbf{75.0} & 57.5 & 5.5 & 3.9 & 41.0
 & 5.5$_{\pm0.5}$ & 47.8$_{\pm1.2}$ & 37.4$_{\pm0.8}$ & 1.4$_{\pm0.5}$ &
2.4$_{\pm0.0}$ & 13.5$_{\pm3.9}$ & 7.0$_{\pm1.2}$ & 16.4$_{\pm0.7}$\\
\textsc{DMA}$^S$ &  27.9 & 87.2 & 15.0 & 23.1 & 17.9 & 36.6 & 45.1 & 53.9 & 44.4 & 73.0 & 35.5 & 2.5 & 4.4 & 35.9 
 & 22.7$_{\pm0.7}$ & 39.9$_{\pm0.7}$ & 56.1$_{\pm0.4}$ & 9.3$_{\pm0.9}$ & 21.6$_{\pm0.0}$ & \textbf{28.2}$_{\pm1.8}$ & 6.5$_{\pm0.8}$ & 26.3$_{\pm0.3}$ \\
\rowcolor{pink!20}
\textbf{\name$^S$} & \textbf{32.6} & 86.8 & 16.5 & \textbf{24.4} & 29.1 & 40.8 & \textbf{52.9} & 59.0 & 48.3 & 74.0 & \textbf{89.0} & 4.0 & \textbf{5.2} & \textbf{43.3}
 & 67.2$_{\pm0.8}$ & \textbf{54.3}$_{\pm1.2}$ & \textbf{65.1}$_{\pm0.5}$ & \textbf{12.5}$_{\pm1.3}$ &
\textbf{31.0}$_{\pm0.0}$ & 25.1$_{\pm1.8}$ & \textbf{8.2}$_{\pm0.7}$ & \textbf{37.6}$_{\pm0.7}$\\
\bottomrule
\end{tabular}
}
\caption{\textbf{\textit{LongBench}} and \textbf{\textit{Helmet}} scores (\%) of \name~compared with all baselines (Full Results). \name$^F$ uses full attention during the prefill stage and sparse attention during the decode stage, while \name$^S$ adopts sparse attention for both stages. See Appendix~\ref{sec:datasets} for \textit{LongBench} task abbreviations.}
\label{tab:results-longbench}
\end{table*}

\begin{table}[t]
\small
\centering
\setlength{\tabcolsep}{10pt}
\resizebox{\linewidth}{!}
{
\begin{tabular}{l|ccccccc|c}
\toprule
{Method} & Recall & RAG & ICL & Cite & Rerank & LongQA & Summ. & Avg. $\uparrow$\\
\midrule

\textsc{FullAttn}
& 21.7$_{\pm0.8}$ 
& 38.1$_{\pm3.9}$ 
& 58.0$_{\pm0.8}$ 
& 2.9$_{\pm0.3}$ 
& 3.3$_{\pm1.0}$ 
& 13.9$_{\pm2.1}$ 
& 4.9$_{\pm1.7}$ 
& 20.4$_{\pm0.6}$ 
\\\midrule

\textsc{InfLLMv2}
& 12.0$_{\pm0.6}$ 
& 30.9$_{\pm3.8}$ 
& 49.0$_{\pm4.6}$ 
& 5.4$_{\pm0.9}$ 
& 0.5$_{\pm0.3}$ 
& 5.0$_{\pm1.4}$ 
& 1.5$_{\pm0.2}$ 
& 14.9$_{\pm0.9}$ 
\\

\textsc{InfLLM}
& 11.2$_{\pm0.8}$ 
& 36.5$_{\pm2.9}$ 
& 35.3$_{\pm3.4}$ 
& \textbf{6.1}$_{\pm1.5}$ 
& 0.6$_{\pm0.4}$ 
& 6.1$_{\pm0.8}$ 
& \textbf{12.3}$_{\pm4.0}$ 
& 15.4$_{\pm0.7}$ 
\\

\textsc{ShadKV}$_{64}$
& 10.0$_{\pm1.4}$ 
& \textbf{41.1}$_{\pm2.5}$ 
& 52.7$_{\pm1.9}$ 
& 1.6$_{\pm0.4}$ 
& 0.0$_{\pm0.0}$ 
& 3.9$_{\pm5.5}$ 
& 2.1$_{\pm0.7}$ 
& 15.9$_{\pm0.8}$ 
\\

\textsc{ShadKV}$_{8}$
& 13.4$_{\pm1.2}$ 
& 38.7$_{\pm3.7}$ 
& 56.7$_{\pm2.1}$ 
& 3.6$_{\pm0.2}$ 
& \textbf{0.8}$_{\pm0.2}$ 
& 4.4$_{\pm2.1}$ 
& 0.5$_{\pm0.7}$ 
& 16.9$_{\pm0.6}$ 
\\

\textsc{ShadKV}$^M_{64}$
& 10.4$_{\pm1.1}$ 
& 32.9$_{\pm5.8}$ 
& 58.3$_{\pm1.3}$ 
& 1.9$_{\pm0.1}$ 
& 0.0$_{\pm0.0}$ 
& \textbf{17.8}$_{\pm11.0}$ 
& 3.9$_{\pm1.4}$ 
& 17.9$_{\pm1.5}$ 
\\

\textsc{ShadKV}$^M_{8}$
& \textbf{14.7}$_{\pm0.7}$ 
& 33.6$_{\pm4.9}$ 
& 63.7$_{\pm4.2}$ 
& 3.0$_{\pm1.1}$ 
& 0.1$_{\pm0.1}$ 
& 11.1$_{\pm6.3}$ 
& 0.0$_{\pm0.0}$ 
& 18.0$_{\pm0.6}$ 
\\

\textsc{DMA}
& 3.1$_{\pm1.9}$ 
& 34.2$_{\pm4.1}$ 
& 54.3$_{\pm3.8}$ 
& 3.0$_{\pm2.3}$ 
& 1.3$_{\pm0.6}$ 
& 1.7$_{\pm2.4}$ 
& 1.5$_{\pm1.2}$ 
& 14.2$_{\pm1.4}$ 
\\

\textsc{DMA}$^S$
& 3.4$_{\pm2.1}$ 
& 29.5$_{\pm4.2}$ 
& 52.3$_{\pm6.7}$ 
& 5.1$_{\pm1.1}$ 
& 0.1$_{\pm0.2}$ 
& 6.1$_{\pm4.2}$ 
& 4.2$_{\pm2.6}$ 
& 14.4$_{\pm1.3}$ 
\\

\rowcolor{pink!20}
\textbf{\textsc{NOSA}}
& 13.2$_{\pm1.2}$ 
& 34.1$_{\pm5.4}$ 
& \textbf{68.3}$_{\pm1.3}$ 
& 3.8$_{\pm0.4}$ 
& 0.7$_{\pm0.2}$ 
& 12.2$_{\pm7.5}$ 
& 10.6$_{\pm6.0}$ 
& \textbf{20.4}$_{\pm1.1}$ 
\\

\bottomrule
\end{tabular}
}
\caption{
\textbf{\textit{HELMET}} benchmark results (\%) at 128K. 
}
\label{tab:helmet-results-128k}
\end{table}

\section{Experiments}

To examine \name's performance and efficiency, we conduct an extensive benchmark to address the following questions: \\
\noindent\:(\underline{\textbf{Q1}}) Does \name~achieve better performance on long-context inputs compared with other baselines? \\
\noindent\:(\underline{\textbf{Q2}}) Can \name~perform competitively on short-context general tasks and long-generation reasoning tasks? \\
\noindent\:(\underline{\textbf{Q3}}) Does \name~deliver higher decoding efficiency?

We also conduct a detailed ablation study in Section~\ref{sec:abl}.

\subsection{Experimental Setup}

We briefly describe the major experimental settings here. Additional details can be found in Appendix~\ref{sec:reproducibility}.

\textbf{Datasets.}
We select representative benchmark datasets covering long-context input, long-generation, general tasks, and efficiency evaluation. 
Please refer to Appendix~\ref{sec:datasets} for tested datasets and Appendix~\ref{sec:benchmark-settings} for detailed settings.

\textbf{Models.} We evaluate three models (1B/3B/8B) which are pretrained at 4K context length then extended to 16K with \name~($k=4096, k_q=1024$) by long-context continual pretraining, followed by SFT. \textsc{InfLLMv2} and \textsc{DMA} use the same pipeline. More details are in Appendix~\ref{sec:model-training}.

\textbf{Baselines.}
\textsc{InfLLMv2} and \textsc{FullAttn} are treated as the upper bound of our method. We include \textsc{ShadowKV}~\citep{sun2024shadowkv} (abbreviated as \textsc{ShadKV}), \textsc{InfLLM}~\citep{xiao2024infllm}, and \textsc{ArkVale}~\citep{chen2024arkvale} as our primary training-free offloading baselines. We also include \textsc{vLLM}~\citep{kwon2023efficient} and \textsc{SGLang}~\citep{zheng2024sglang} in the efficiency benchmark.
Please refer to Appendix~\ref{sec:method-params} and~\ref{sec:baselines} for detailed settings and descriptions.

\subsection{Long Context Tasks}
\label{sec:long-context-tasks}

To answer \underline{\textbf{Q1}}, we evaluate \name~against the baselines on LongBench and HELMET (Table~\ref{tab:results-longbench}).


\textbf{\name~achieves the best overall performance among all competing methods.} Across the 12 settings in Table~\ref{tab:results-longbench}, \name\ attains the highest score in 9 settings, while \textsc{ShadowKV} and \textsc{ArkVale} perform best in 2 and 1 settings, respectively. Overall, \name\ outperforms the other methods in most cases and achieves the largest number of wins.

\textbf{\name~excels in hard context retrieval tasks.}
For the baselines, introducing sparsity patterns unseen during training severely degrades their ability to handle hard retrieval tasks. For example, all baselines suffer a substantial performance drop ($\geq 10\%$) on the PR task in LongBench (sparse prefill) and the Recall task in HELMET (full prefill). In contrast, \name~maintains consistent performance and exhibits only a modest degradation compared to other baselines. 

Notably, due to \textsc{DMA}'s inability to recall evicted context by design, it exhibits low performance on both overall and recall tasks.

\textbf{\name~achieves consistently good performance at longer input lengths.} To evaluate whether \name~can achieve better performance under long-context extension, we evaluate the 8B model on HELMET at 128K context length. Results in Table~\ref{tab:helmet-results-128k} show that while other baselines degrade severely compared with full attention, \name~can still achieve near-lossless performance and outperform all the baselines.

\textbf{\name~benefits from larger models.} Sparse attention methods tend to be less stable on smaller models, which are generally less robust than larger ones. For example, on the 1B model, applying \textsc{InfLLMv2} alone leads to a recall drop of more than 20\% on HELMET, and all sparse attention methods exhibit notable performance degradation on HELMET at this scale. In contrast, for larger models (8B), \name~consistently achieves the highest scores among all competing methods, and the degradation in recall performance is substantially less severe than that observed on smaller models. These results indicate that sparse attention methods, particularly \name, benefit from increased model capacity and larger-scale training data, which together yield a more robust base model.

\begin{table}[t]
\small
\noindent

\begin{minipage}[t]{0.45\linewidth}
\centering
\setlength{\tabcolsep}{2.5pt}

\scalebox{0.74}{
\begin{tabular}{ll|cccccccc|c}
\toprule
Model & Method & MM & MMP & BB & GS & MA & DR & MB & HE & Avg. $\uparrow$\\
\midrule
\multirow{4}{*}{\textit{1B}} &
\textsc{FullAttn} & 50.3 & 24.2 & 37.8 & 48.5 & 12.0 & 8.6 & 42.4 & 43.3 & 33.4\\
&\textsc{InfLLMv2} & 48.5 & 24.2 & 37.6 & 50.9 & 11.7 & 8.6 & 40.0 & 39.0 & 32.6 \\
&\textsc{DMA} & 48.8 & 23.8 & 37.8 & 51.9 & 11.6 & 7.2 & 40.0 & 43.3 & 33.0 \\
& \cellcolor{pink!20}\name
& \cellcolor{pink!20}48.8
& \cellcolor{pink!20}23.8
& \cellcolor{pink!20}39.1
& \cellcolor{pink!20}51.0
& \cellcolor{pink!20}11.6
& \cellcolor{pink!20}7.2
& \cellcolor{pink!20}40.2
& \cellcolor{pink!20}45.7
& \cellcolor{pink!20}33.4 \\
\midrule
\multirow{4}{*}{\textit{3B}}
&\textsc{FullAttn} & 60.1 & 34.8 & 54.2 & 68.5 & 17.4 & 8.6 & 51.6 & 56.7 & 44.0\\
&\textsc{InfLLMv2} &  60.4 & 35.2 & 51.4 & 68.8 & 17.7 & 8.6 & 51.6 & 55.5 & 43.6 \\
&\textsc{DMA} & 60.0 & 35.8 & 52.8 & 67.3 & 17.2 & 8.3 & 53.2 & 54.3 & 43.7\\
& \cellcolor{pink!20}\name
& \cellcolor{pink!20}59.9
& \cellcolor{pink!20}34.8
& \cellcolor{pink!20}53.4
& \cellcolor{pink!20}68.2
& \cellcolor{pink!20}17.5
& \cellcolor{pink!20}8.5
& \cellcolor{pink!20}51.8
& \cellcolor{pink!20}57.3
& \cellcolor{pink!20}43.9\\
\midrule
\multirow{4}{*}{\textit{8B}}
&\textsc{FullAttn} & 73.6 & 46.1 & 61.9 & 76.3 & 22.2 & 11.3 & 62.6 & 69.5 & 52.9  \\
&\textsc{InfLLMv2} & 73.2 & 45.6 & 63.1 & 77.1 & 22.2 & 10.3 & 63.0 & 67.7 & 52.8 \\
&\textsc{DMA} & 69.8 & 45.5 & 69.4 & 76.7 & 22.3 & 11.1 & 62.2 & 69.5 & 53.3 \\
& \cellcolor{pink!20}\name
& \cellcolor{pink!20}69.8
& \cellcolor{pink!20}45.8
& \cellcolor{pink!20}68.9
& \cellcolor{pink!20}77.8
& \cellcolor{pink!20}22.3
& \cellcolor{pink!20}10.6
& \cellcolor{pink!20}62.6
& \cellcolor{pink!20}71.3
& \cellcolor{pink!20}53.6 \\
\bottomrule
\end{tabular}
}

\caption{The performance (\%) on \textbf{\textit{general tasks}}. All models are evaluated with full attention. Abbreviations are in Appendix~\ref{sec:datasets}.}
\label{tab:results-general}
\end{minipage}
\hspace{7pt}
\begin{minipage}[t]{0.48\linewidth}
\centering
\setlength{\tabcolsep}{1pt}

\scalebox{0.74}{
\begin{tabular}{l|cc|cccc>{\columncolor{pink!20}}c}
\toprule
Dataset & \textsc{FullAttn} & \textsc{InfLLMv2} & \textsc{ShadKV} & \textsc{InfLLM} & \textsc{ArkVale} & \textsc{DMA} & \textbf{\name}\\
\midrule
Math-500 & 52.8 & 50.6 & 19.0 & 38.8 & 47.0 & \textbf{50.4} & \textbf{50.4} \\
Gaokao-MS & 55.6 & 49.5 & 7.0 & 36.9 & 43.9 & \textbf{57.0} & 51.9 \\
Gaokao-MH & 61.5 & 57.8 & 15.6 & 42.2 & 47.7 & 58.7 & \textbf{62.4}\\
Gaokao-Phy & 36.3 & 34.8 & 8.2 & 29.7 & 30.5 & 31.6 & \textbf{33.6}\\
\midrule
Avg. $\uparrow$ & 51.6 & 48.2 & 12.5 & 36.9 & 42.3 & 49.4 & \textbf{49.6}\\
\bottomrule
\end{tabular}
}

\caption{The performance (\%) on \textbf{\textit{reasoning tasks}}. 
All experiments are conducted on the 8B model.}
\label{tab:results-reasoning}
\end{minipage}
\vspace{-10pt}
\end{table}

\subsection{General and Reasoning Tasks}

To address \underline{\textbf{Q2}}, we evaluate general tasks (Table~\ref{tab:results-general}) and long-generation reasoning tasks (Table~\ref{tab:results-reasoning}). 

\textbf{\name~does not harm short-context general ability.}
As shown in Table~\ref{tab:results-general}, all methods achieve comparable performance on the general tasks across model sizes. Both \name~and \textsc{DMA} exhibit no significant performance degradation compared with \textsc{InfLLMv2} and \textsc{FullAttn}. 

\textbf{Training-free offloading methods fail on long-generation reasoning tasks.}
Table~\ref{tab:results-reasoning} shows that training-free offloading methods incur substantial performance drops on reasoning tasks. In contrast, \textsc{DMA} and \name~perform better, as their sparsity patterns are learned during training.
To provide more insight, we conduct a case study and measure perplexity over one generation in Figures~\ref{fig:ppl}. Training-free offloading (\textsc{ShadowKV}) shows sharply increasing perplexity, whereas \name's perplexity remains low, indicating that training-inference mismatch accumulates error and corrupts reasoning generations (Section~\ref{sec:case-study}).

\begin{table*}[t]
\small

\centering
\setlength{\tabcolsep}{0.8pt} 

\scalebox{0.7}{
\begin{tabular}{llc|cccc!{\vrule width \heavyrulewidth}llc|cccc}
\toprule
\multirow{2}{*}{Method} & \multirow{2}{*}{Impl.} & \multirow{2}{*}{Off.} & \multicolumn{4}{c!{\vrule width \heavyrulewidth}}{\textbf{\textit{Input Length: 16K}}} & \multirow{2}{*}{Method} & \multirow{2}{*}{Impl.} & \multirow{2}{*}{Off.} & \multicolumn{4}{c}{\textbf{\textit{Input Length: 32K}}} \\
 & & &  \textit{EB: 16} & \textit{EB: 32} & \textit{EB: 64} & \textit{EB: 128} & & & & \textit{EB: 16} & \textit{EB: 32} & \textit{EB: 64} & \textit{EB: 128} \\
\midrule
\textsc{FullAttn} & \textsc{HF} & \ding{55} & 160.85 (4) & 239.94 (8) & OOM (16) & OOM (32) & \textsc{FullAttn} & \textsc{HF} & \ding{55} & 80.56 (2) & 120.99 (4) & OOM (8) & OOM (16)
\\
\textsc{FullAttn} & \textsc{vLLM} & \ding{55} & 233.21 (4)  & 417.99 (8) & 678.41 (16) & 927.99 (32) & \textsc{FullAttn} & \textsc{vLLM} & \ding{55} & 113.42 (2) & 151.58 (4) & 460.65 (8)  & 818.53 (16)
\\
\textsc{FullAttn} & \textsc{SGLang} & \ding{55} & 180.28 (4) & 358.56 (8) & 774.74 (16) & 1267.20  (32) & \textsc{FullAttn} & \textsc{SGLang} & \ding{55} & 116.54 (2) & 174.50 (4) & 290.29 (8)  & 748.20 (16)
\\
\midrule
\textsc{InfLLMv2} &\textsc{HF} & \ding{55} & 47.10 (4) & OOM (8) & OOM (16) & OOM (32) & \textsc{InfLLMv2} & \textsc{HS} & \ding{55} & 23.17 (2) & OOM (4) & OOM (8) & OOM (16) 
\\
\textsc{InfLLMv2} & \namei & \ding{55} & 226.51 (4) & 440.76 (8) & 824.09 (16)  & 1475.43 (32) & \textsc{InfLLMv2} & \namei & \ding{55} & 104.88 (2) & 205.74 (4) & 403.26 (8) & 710.93 (16)
\\
\textsc{InfLLMv2} & \namei & \ding{51} & 563.88 (16) & 759.35 (32) & 1389.40 (64) & 1441.32 (128) & \textsc{InfLLMv2} & \namei & \ding{51} & 491.42 (16) & 623.50 (32) & 1132.50 (64) & 652.60 (128)
\\
\midrule
\textsc{ShadKV} & \textsc{ShadKV} & \ding{51} & 585.92 (16) & 769.84 (32) & 970.19 (64)  & 1071.36 (128) & \textsc{ShadKV} & \textsc{ShadKV} & \ding{51} & 541.91 (16) & 697.10 (32) & 885.05 (64) & 995.81 (128)
\\
\textsc{ArkVale} & \textsc{ArkVale} & \ding{51} & 34.15 (1) & 46.76 (2)  & 57.38 (4) & 68.65 (8) & \textsc{ArkVale} & \textsc{ArkVale} & \ding{51} & 29.45 (1) & 38.15 (2) & 45.05 (4) & OOM (8)
\\
\textsc{InfLLM} & \textsc{InfLLM}  & \ding{51} & 35.00 (16) & OOM (32) & OOM (64) & OOM (128) & \textsc{InfLLM} & \textsc{InfLLM} & \ding{51} & 30.72 (16) & OOM (32) & OOM (64) & OOM (128)
\\
\midrule
\name & \textsc{HF}  & \ding{55} & 14.69 (4) & OOM (8) & OOM (16) & OOM (32) & \name & \textsc{HF} & \ding{55} & 3.62 (2) & OOM (4) & OOM (8) & OOM (16)
\\
\name & \namei  & \ding{55} & 206.10 (4) & 411.72 (8) & 817.12 (16) & 1503.76 (32) & \name & \namei & \ding{55} & 96.15 (2) & 205.18 (4) & 402.82 (8) & 772.19 (16)
\\
\rowcolor{pink!20}
\textbf{\name} & \textbf{\namei}  & \ding{51} & \textbf{743.18} (16) & \textbf{1056.27} (32) & \textbf{1630.80} (64) & \textbf{1961.17} (128) & \textbf{\name} & \textbf{\namei} & \ding{51} & \textbf{694.58} (16) & \textbf{1067.32} (32) & \textbf{1574.47} (64) & \textbf{1536.19} (128)
\\

\specialrule{\heavyrulewidth}{0.5ex}{0.5ex}


\multirow{2}{*}{Method} & \multirow{2}{*}{Impl.}  & \multirow{2}{*}{Off.} & \multicolumn{4}{c!{\vrule width \heavyrulewidth}}{\textbf{\textit{Input Length: 64K}}} & \multirow{2}{*}{Method} & \multirow{2}{*}{Impl.} & \multirow{2}{*}{Off.} & \multicolumn{4}{c}{\textbf{\textit{Input Length: 96K}}} \\
 & & &  \textit{EB: 8} & \textit{EB: 16} & \textit{EB: 32} & \textit{EB: 64} & & & & \textit{EB: 8} & \textit{EB: 16} & \textit{EB: 32} & \textit{EB: 64}  \\
\midrule
\textsc{FullAttn} & \textsc{HF} & \ding{55} & -- & 40.80 (1) & 60.41 (2) & OOM (4) & \textsc{FullAttn} & \textsc{HF} & \ding{55} & -- & 35.17 (1) & 48.67 (2) & OOM (4)
\\
\textsc{FullAttn} & \textsc{vLLM} & \ding{55} & -- & 60.22 (1) & 82.40 (2) & 240.28 (4) & \textsc{FullAttn} & \textsc{vLLM} & \ding{55} & -- & 48.48 (1) & 120.76 (2) & 206.90 (4)
\\
\textsc{FullAttn} & \textsc{SGLang} & \ding{55} & -- & 45.19 (1) & 81.60 (2) & 191.65 (4) & \textsc{FullAttn} & \textsc{SGLang} & \ding{55} & -- & 47.23 (1) & 78.47 (2) & 214.42 (4)
\\
\midrule
\textsc{InfLLMv2} &\textsc{HF} & \ding{55} & -- & 11.14 (1) & OOM (2) & OOM (4) & \textsc{InfLLMv2} & \textsc{HS} & \ding{55} & -- & 9.26 (1) & OOM (2) & OOM (4) 
\\
\textsc{InfLLMv2} & \namei & \ding{55} & -- & 46.62 (1) & 89.77 (2) & 174.66 (4) & \textsc{InfLLMv2} & \namei & \ding{55} & -- & 40.05 (1) & 78.06 (2) & 153.15 (4)
\\
\textsc{InfLLMv2} & \namei & \ding{51} &  273.54 (8) & 409.34 (16) & 561.95 (32) & 717.65 (64)  & \textsc{InfLLMv2} & \namei & \ding{51} & 264.92 (8) & 506.80 (16) & 444.61 (32) & 725.26 (64)
\\ 
\midrule
\textsc{ShadKV} & \textsc{ShadKV} & \ding{51} & 292.97 (8) & 503.73 (16) & 686.69 (32) & 866.01 (64) & \textsc{ShadKV} & \textsc{ShadKV} & \ding{51} & 271.40 (8) & 457.08 (16) & 625.71 (32) & 730.48 (64)
\\
\textsc{ArkVale} & \textsc{ArkVale} & \ding{51} & -- & 25.42 (1) & 32.83 (2) & OOM (4) & \textsc{ArkVale} & \textsc{ArkVale} & \ding{51} & -- & 24.43 (1) & 30.35 (2) & OOM (4)
\\
\textsc{InfLLM} & \textsc{InfLLM}  & \ding{51} & 30.04 (8) & OOM (16) & OOM (32) & OOM (64) & \textsc{InfLLM} & \textsc{InfLLM} & \ding{51} & 24.96 (8) & OOM (16) & OOM (32) &  OOM (64)
\\
\midrule
\name & \textsc{HF}  & \ding{55} & -- & 0.81 (1) & OOM (2) & OOM (4) & \name & \textsc{HF} & \ding{55} & -- & 0.27 (1) & OOM (2) & OOM (4) 
\\
\name & \namei  & \ding{55} & -- & 46.08 (1) & 84.91 (2) & 176.23 (4) & \name & \namei & \ding{55} & -- & 38.08 (1) & 78.50 (2) & 154.11 (4) 
\\
\rowcolor{pink!20}
\textbf{\name} & \textbf{\namei}  & \ding{51} & \textbf{312.39} (8) & \textbf{545.08} (16) & \textbf{763.82} (32) & \textbf{1378.82} (64) & \textbf{\name} & \textbf{\namei} & \ding{51} &  \textbf{283.44} (8) & \textbf{542.03} (16) & \textbf{666.59} (32) & \textbf{1080.45} (64)
\\

\bottomrule
\end{tabular}
}
\caption{
Decoding throughput (tok/s) $\uparrow$. ``EB'' = equivalent batch size; ``Impl.”/“Off.'' = implementation/offloading. For each EB, we fix GPU KV cache size and report the real batch size in parentheses. ``--'' indicates no feasible batch ($\ge1$); ``OOM'' = out-of-memory. 
}
\label{tab:results-thru}
\vspace{-8pt}
\end{table*}

\subsection{Efficiency Benchmarks}

To address \underline{\textbf{Q3}}, we compare \name~with baseline methods in terms of decoding throughput and evaluate different implementations to demonstrate the effectiveness of \namei.

\textbf{Larger batch sizes lead to higher decoding efficiency.} Table~\ref{tab:results-thru} shows that offloading methods (\textsc{ShadowKV}, \name) achieve higher decoding throughput than non-offloading baselines, because the memory constraint allows a larger achievable batch size. \name~achieves up to 5.04$\times$ higher decoding throughput (input length 96K, EB: 64) than \textsc{FullAttn}, since the real batch size is 16$\times$ larger when the on-GPU KV cache sizes are matched.

\textbf{\name~achieves the highest decoding throughput.} Among offloading-based approaches, \name~consistently delivers the best decoding throughput. \textsc{ArkVale} does not utilize grouped-query attention (GQA)\footnote{\textsc{ArkVale} only supports GQA with a $q$:$k$ ratio of 1:1, 4:1, or 8:1. For our model (16:1), the official impl. degrades to MHA.} for our model architecture, resulting in a smaller achievable batch size and consequently limited throughput. \textsc{InfLLM} suffers from an inefficient implementation, leading to low decoding throughput and high GPU memory utilization. \textsc{ShadowKV} is the most competitive baseline; nevertheless, \name~achieves up to 1.83$\times$ higher throughput than \textsc{ShadowKV}.

\textbf{Higher locality leads to faster decoding speed.}
Since \name~explicitly incorporates locality constraints, it achieves better locality and less CPU-GPU communication than \textsc{InfLLMv2}. Consequently, \name~attains up to a $1.92\times$ higher decoding throughput over \textsc{InfLLMv2}.

\textbf{Implementation matters.}\footnote{We use the original \texttt{generate} for \textsc{HF} (may OOM due to large hidden states). Decoding time in \textsc{vLLM} is measured from the first to the last generated token. For \textsc{SGLang}, we estimate it as (prefill+decode) latency minus prefill-only latency.} \namei~achieves substantially higher decoding throughput than the vanilla HF implementation. Under HF, \name~is slower than \textsc{InfLLMv2} due to suboptimal kernels and excessive small kernel launches. However, under \namei, the overheads are largely mitigated, revealing communication as the true bottleneck; consequently, \name~surpasses all baselines.

\subsection{Ablation Studies}
\label{sec:abl}


To further demonstrate \name’s performance and efficiency, we conduct comprehensive ablation studies on length generalization, hyperparameter sensitivity, efficiency under Various Semantics, various hardware configurations, and component-wise breakdown, followed by a case study on long-generation.

\begin{table*}[t]
\small

\centering
\setlength{\tabcolsep}{1.8pt} 
\scalebox{0.74}{
\begin{tabular}{l|c|cccc>{\columncolor{pink!20}}c!{\vrule width \heavyrulewidth}c|ccccc>{\columncolor{pink!20}}c}
\toprule
\multirow{2}{*}{Tasks} & \multicolumn{6}{c!{\vrule width \heavyrulewidth}}{\textit{Full Prefill}} & \multicolumn{7}{c}{\textit{Sparse Prefill}} \\
 & \textsc{FullAttn} & \textsc{ShadKV}$_{64}$ & \textsc{ShadKV}$_8$ & \textsc{ArkVale} & \textsc{DMA}$^F$ & \textbf{\name}$^F$ & \textsc{InfLLMv2} & \textsc{ShadKV}$_{64}^M$ & \textsc{ShadKV}$_8^M$ & \textsc{InfLLM}$_{128}$ & \textsc{InfLLM}$_{64}$ & \textsc{DMA}$^S$ & \textbf{\name}$^S$ \\
\midrule
\multicolumn{14}{c}{\small\textbf{\textit{Budget $k=2048$ \quad Sparsity: 0.125}}} \\
\midrule
Recall &
88.0$_{\pm0.8}$ & 40.3$_{\pm4.3}$ & 72.6$_{\pm1.1}$ & 49.3$_{\pm3.8}$ & 17.5$_{\pm1.1}$ & \textbf{73.9}$_{\pm0.8}$ & 54.2$_{\pm0.9}$ & 30.5$_{\pm3.0}$ & \textbf{56.0}$_{\pm2.8}$ & 20.3$_{\pm2.3}$ & 20.4$_{\pm2.2}$ & 10.7$_{\pm2.1}$ & 40.4$_{\pm2.5}$ 
\\
RAG & 
54.5$_{\pm1.9}$ & 53.3$_{\pm2.5}$ & 53.8$_{\pm2.2}$ & 54.0$_{\pm2.6}$ & 52.2$_{\pm3.2}$ & \textbf{56.4}$_{\pm2.0}$ & 51.8$_{\pm2.3}$ & 44.5$_{\pm2.9}$ & 43.6$_{\pm1.4}$ & 46.3$_{\pm3.6}$ & 46.7$_{\pm2.2}$ & 28.7$_{\pm1.8}$ & \textbf{46.8}$_{\pm1.4}$
\\
ICL & 
59.7$_{\pm2.1}$ & 55.7$_{\pm0.5}$ & 57.3$_{\pm2.1}$ & 60.7$_{\pm2.6}$ & \textbf{63.7}$_{\pm2.5}$ & 59.0$_{\pm2.4}$ & 66.3$_{\pm3.3}$ & 52.3$_{\pm2.6}$ & 49.3$_{\pm5.3}$ & 41.3$_{\pm3.4}$ & 42.7$_{\pm2.1}$ & 42.3$_{\pm3.9}$ & \textbf{60.7}$_{\pm5.4}$
\\
Cite & 
13.9$_{\pm3.7}$ & 9.7$_{\pm0.6}$ & 8.6$_{\pm0.5}$ & 10.9$_{\pm1.1}$ & 9.2$_{\pm2.8}$ & \textbf{11.9}$_{\pm2.3}$ & 12.7$_{\pm0.9}$ & 4.9$_{\pm0.9}$ & 6.9$_{\pm1.5}$ & 8.3$_{\pm0.8}$ & 9.6$_{\pm1.2}$ & 7.8$_{\pm2.9}$ & \textbf{12.7}$_{\pm1.0}$
\\
Rerank & 
36.4$_{\pm3.1}$ & 13.3$_{\pm1.6}$ & 26.5$_{\pm2.3}$ & \textbf{29.2}$_{\pm2.5}$ & 17.4$_{\pm2.2}$ & 24.7$_{\pm2.6}$ & 26.4$_{\pm1.7}$ & 6.0$_{\pm0.7}$ & 13.1$_{\pm1.2}$ & 8.1$_{\pm0.5}$ & 8.2$_{\pm0.9}$ & 8.0$_{\pm1.6}$ & \textbf{22.6}$_{\pm1.8}$
\\
LongQA & 
31.0$_{\pm2.2}$ & 27.9$_{\pm4.2}$ & 32.6$_{\pm4.3}$ & 28.8$_{\pm1.2}$ & 29.5$_{\pm3.4}$ & \textbf{31.2}$_{\pm1.3}$ & 28.5$_{\pm3.6}$ & 25.0$_{\pm2.3}$ & 25.5$_{\pm2.9}$ & 9.2$_{\pm4.0}$ & 12.5$_{\pm4.2}$ & \textbf{29.6}$_{\pm2.6}$ & 28.6$_{\pm4.6}$
\\
Summ. & 
11.3$_{\pm4.8}$ & \textbf{9.8}$_{\pm4.8}$ & 7.0$_{\pm3.2}$ & 7.6$_{\pm2.0}$ & 4.2$_{\pm0.6}$ & 7.4$_{\pm3.7}$ & 9.1$_{\pm3.2}$ & 4.1$_{\pm0.8}$ & 3.9$_{\pm2.5}$ & 8.2$_{\pm3.8}$ & \textbf{9.1}$_{\pm4.5}$ & 6.7$_{\pm2.7}$ & 7.0$_{\pm1.3}$
\\
\midrule
Avg. $\uparrow$ & 
42.1$_{\pm0.8}$ & 30.0$_{\pm1.2}$ & 36.9$_{\pm1.2}$ & 34.3$_{\pm0.9}$ & 27.7$_{\pm0.2}$ & \textbf{37.8}$_{\pm0.4}$ & 35.6$_{\pm0.8}$ & 23.9$_{\pm0.5}$ & 28.3$_{\pm1.1}$ & 20.2$_{\pm0.7}$ & 21.3$_{\pm0.4}$ & 19.1$_{\pm1.0}$ & \textbf{31.3}$_{\pm1.9}$
\\
\midrule
\multicolumn{14}{c}{\small\textbf{\textit{Budget $k=4096$ \quad Sparsity: 0.250}}} \\
\midrule
Recall &
88.0$_{\pm0.8}$ & 56.3$_{\pm3.6}$ & 78.5$_{\pm1.1}$ & 58.6$_{\pm1.6}$ & 31.9$_{\pm4.9}$ & \textbf{85.5}$_{\pm1.2}$ & 74.2$_{\pm3.6}$ & 55.7$_{\pm6.2}$ & \textbf{70.5}$_{\pm3.0}$ & 26.8$_{\pm1.3}$ & 32.4$_{\pm2.7}$ & 20.9$_{\pm1.0}$ & 67.2$_{\pm1.3}$ 
\\
RAG & 
54.5$_{\pm1.9}$ & 53.8$_{\pm1.7}$ & 53.1$_{\pm2.2}$ & 54.3$_{\pm2.1}$ & 53.1$_{\pm2.8}$ & \textbf{56.2}$_{\pm3.0}$ & 53.1$_{\pm2.0}$ & \textbf{53.4}$_{\pm1.8}$ & 51.8$_{\pm1.6}$ & 46.3$_{\pm2.1}$ & 46.4$_{\pm2.4}$ & 40.0$_{\pm2.5}$ & 52.6$_{\pm1.7}$ 
\\
ICL & 
59.7$_{\pm2.1}$ & 57.3$_{\pm3.9}$ & 58.3$_{\pm0.5}$ & 61.3$_{\pm2.6}$ & \textbf{67.3}$_{\pm0.9}$ & 61.3$_{\pm1.7}$ & 72.0$_{\pm0.8}$ & 53.7$_{\pm3.4}$ & 55.0$_{\pm1.4}$ & 39.0$_{\pm2.2}$ & 38.0$_{\pm3.7}$ & 59.3$_{\pm1.2}$ & \textbf{67.0}$_{\pm2.2}$ 
\\
Cite & 
13.9$_{\pm3.7}$ & 9.6$_{\pm2.6}$ & 8.9$_{\pm2.3}$ & 11.6$_{\pm1.3}$ & 9.9$_{\pm0.5}$ & \textbf{12.4}$_{\pm0.5}$ & 12.8$_{\pm3.5}$ & 9.7$_{\pm1.2}$ & 7.9$_{\pm2.3}$ & \textbf{12.8}$_{\pm1.4}$ & 11.2$_{\pm1.6}$ & 9.6$_{\pm1.5}$ & 10.4$_{\pm0.4}$ 
\\
Rerank & 
36.4$_{\pm3.1}$ & 23.7$_{\pm1.7}$ & 28.6$_{\pm2.3}$ & 32.8$_{\pm2.6}$ & 25.8$_{\pm3.8}$ & \textbf{33.4}$_{\pm0.9}$ & 35.5$_{\pm2.9}$ & 20.8$_{\pm3.5}$ & 25.5$_{\pm3.6}$ & 12.4$_{\pm0.5}$ & 14.1$_{\pm0.2}$ & 17.3$_{\pm1.8}$ & \textbf{30.4}$_{\pm2.7}$ 
\\
LongQA & 
31.0$_{\pm2.2}$ & 29.2$_{\pm4.4}$ & \textbf{31.9}$_{\pm6.3}$ & 30.8$_{\pm1.9}$ & 30.7$_{\pm2.5}$ & 31.4$_{\pm1.7}$ & 31.4$_{\pm1.6}$ & 26.4$_{\pm3.4}$ & \textbf{31.9}$_{\pm7.0}$  & 11.6$_{\pm3.5}$ & 11.1$_{\pm1.9}$ & 31.5$_{\pm4.6}$ & {28.5}$_{\pm3.6}$ 
\\
Summ. & 
11.3$_{\pm4.8}$ & 9.9$_{\pm4.2}$ & \textbf{10.4}$_{\pm3.4}$ & 8.9$_{\pm3.5}$ & 8.3$_{\pm2.6}$ & 9.1$_{\pm2.4}$ & 11.2$_{\pm3.6}$ & \textbf{11.1}$_{\pm3.0}$ & 8.7$_{\pm3.2}$ & 9.1$_{\pm2.5}$ & {10.0}$_{\pm5.8}$ & 9.4$_{\pm4.3}$ & 8.9$_{\pm2.3}$ 
\\
\midrule
Avg. $\uparrow$ & 
42.1$_{\pm0.8}$ & 34.3$_{\pm0.5}$ & 38.5$_{\pm1.7}$ & 36.9$_{\pm0.5}$ & 32.4$_{\pm0.6}$ & \textbf{41.3}$_{\pm0.4}$ & 41.5$_{\pm0.8}$ & 33.0$_{\pm1.1}$ & 35.9$_{\pm0.1}$  & 22.6$_{\pm0.4}$ & 23.3$_{\pm1.6}$ & 26.9$_{\pm0.6}$ & \textbf{37.9}$_{\pm0.8}$ 
\\
\midrule
\multicolumn{14}{c}{\small\textbf{\textit{Budget $k=6144$ \quad Sparsity: 0.375}}} \\
\midrule
Recall &
88.0$_{\pm0.8}$ & 68.0$_{\pm2.0}$ & 79.7$_{\pm1.6}$ & 68.0$_{\pm3.0}$ & 46.2$_{\pm2.9}$ & \textbf{93.6}$_{\pm2.2}$ & 86.7$_{\pm1.6}$ & 61.0$_{\pm4.9}$ & 71.4$_{\pm3.4}$ & 34.4$_{\pm1.4}$ & 38.1$_{\pm3.1}$ & 36.1$_{\pm2.1}$ & \textbf{83.1}$_{\pm4.6}$ 
\\
RAG & 
54.5$_{\pm1.9}$ & 54.0$_{\pm2.0}$ & 53.0$_{\pm2.6}$ & 55.1$_{\pm2.4}$ & 53.5$_{\pm3.0}$ & \textbf{56.4}$_{\pm1.7}$ & 55.1$_{\pm2.0}$ & 53.3$_{\pm2.1}$ & 51.9$_{\pm1.3}$ & 46.3$_{\pm1.7}$ & 46.3$_{\pm2.9}$ & 47.6$_{\pm1.4}$ & \textbf{53.6}$_{\pm0.9}$ 
\\
ICL & 
59.7$_{\pm2.1}$ & 58.3$_{\pm2.4}$ & 58.7$_{\pm0.9}$ & 61.0$_{\pm2.2}$ & \textbf{67.3}$_{\pm0.5}$ & 63.0$_{\pm2.2}$ & 68.0$_{\pm1.4}$ & 55.0$_{\pm2.2}$ & 55.3$_{\pm1.9}$ & 40.3$_{\pm2.1}$ & 41.3$_{\pm0.5}$ & 65.3$_{\pm0.9}$ & \textbf{66.7}$_{\pm2.6}$ 
\\
Cite & 
13.9$_{\pm3.7}$ & 10.6$_{\pm0.9}$ & 10.5$_{\pm1.2}$ & 10.5$_{\pm0.9}$ & \textbf{10.7}$_{\pm1.3}$ & \textbf{10.7}$_{\pm1.3}$ & 14.5$_{\pm4.4}$ & 8.8$_{\pm1.4}$ & 8.2$_{\pm1.7}$ & \textbf{12.1}$_{\pm1.5}$ & 11.4$_{\pm1.9}$ & 10.6$_{\pm1.1}$ & 12.0$_{\pm1.5}$ 
\\
Rerank & 
36.4$_{\pm3.1}$ & 25.4$_{\pm2.1}$ & 30.1$_{\pm2.2}$ & 33.5$_{\pm2.1}$ & 30.4$_{\pm1.6}$ & \textbf{37.0}$_{\pm2.6}$ & 39.7$_{\pm1.0}$ & 25.0$_{\pm1.8}$ & 26.4$_{\pm4.0}$ & 18.2$_{\pm1.7}$ & 20.5$_{\pm0.3}$ & 26.3$_{\pm2.0}$ & \textbf{36.9}$_{\pm1.3}$ 
\\
LongQA & 
31.0$_{\pm2.2}$ & 29.7$_{\pm4.2}$ & 31.2$_{\pm6.2}$ & 30.5$_{\pm0.8}$ & \textbf{32.5}$_{\pm3.6}$ & 31.8$_{\pm0.9}$ & 30.2$_{\pm1.8}$ & 26.1$_{\pm4.3}$ & \textbf{32.1}$_{\pm7.7}$ & 11.4$_{\pm2.3}$ & 10.6$_{\pm1.4}$ & 31.1$_{\pm5.5}$ & 31.3$_{\pm2.4}$ 
\\
Summ. & 
11.3$_{\pm4.8}$ & 9.0$_{\pm4.0}$ & 7.8$_{\pm2.5}$ & 10.0$_{\pm2.8}$ & 7.7$_{\pm1.0}$ & \textbf{10.2}$_{\pm2.7}$ & 9.9$_{\pm4.7}$ & 9.9$_{\pm4.4}$ & 7.2$_{\pm2.7}$ & 10.0$_{\pm2.0}$ & 9.8$_{\pm4.9}$ & 10.8$_{\pm2.9}$ & \textbf{11.5}$_{\pm1.9}$ 
\\
\midrule
Avg. $\uparrow$ & 
42.1$_{\pm0.8}$ & 36.4$_{\pm0.5}$ & 38.7$_{\pm1.5}$ & 38.4$_{\pm0.8}$ & 35.5$_{\pm0.6}$ & \textbf{43.3}$_{\pm0.5}$ & 43.4$_{\pm0.8}$ & 34.2$_{\pm0.3}$ & 36.1$_{\pm0.1}$ & 24.7$_{\pm0.4}$ & 25.5$_{\pm0.7}$ & 32.6$_{\pm1.1}$ & \textbf{42.1}$_{\pm1.0}$ 
\\
\bottomrule
\end{tabular}
}
\caption{\textbf{\textit{Helmet}} scores (\%) of \name~compared with all baselines under various budget $k$. \name$^F$ uses full attention during the prefill stage and sparse attention during the decode stage, while \name$^S$ adopts sparse attention for both stages.}
\label{tab:results-budget}
\end{table*}

\subsubsection{Length Generalization.}

To compare \name~with other baselines on longer input lengths and to test its ability to generalize to out-of-domain (OOD) context lengths, we evaluate all methods using the 8B model on HELMET with 32K and 64K inputs. Note that our models are trained with a 16K context window, so these evaluations correspond to $2\times$ and $4\times$ length generalization beyond the training length. For the 32K setting, we use the same protocol as in Section~\ref{sec:long-context-tasks}. For 64K, we increase RoPE's $\theta$ from 10000 to 40000 to improve performance, and increase the sparsity budget (i.e., $k$) to 6144. The results are reported in Table~\ref{tab:results-longer-inputs}, from which we draw the following conclusion.

\textbf{\name~also shows superior performance in length extrapolation.} As shown in Table~\ref{tab:results-longer-inputs}, \name~outperforms all other baselines at both 32K and 64K input lengths. Even when evaluated on OOD context lengths, \name~maintains strong performance compared to alternative methods. These results further demonstrate that \name~is compatible with length extrapolation, as increasing RoPE's $\theta$ is a common practice for extending the context length without retraining, and \name~still performs better under this modification.

\subsubsection{Hyperparameter Analysis}

\begin{wrapfigure}{r}{0.35\textwidth}
  \centering
  \vspace{-20pt}
  \includegraphics[width=\linewidth]{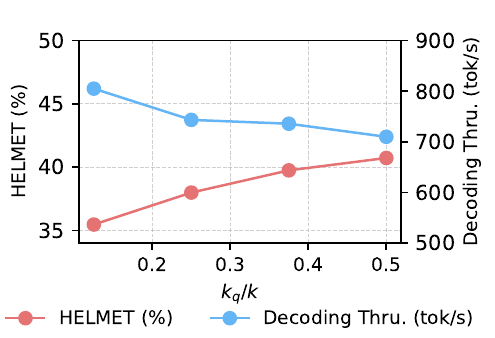}
  \vspace{-10pt}
  \caption{
    HELMET score and decoding throughput under various $k_q/k$.
  }
  \vspace{-15pt}
  \label{fig:sbabl}
\end{wrapfigure}

\textbf{Varying the ratio of query-aware selection $k_q/k$.}
The minimum locality, $\gamma_0 = k_e/k = 1 - k_q/k$, is a crucial hyperparameter that controls the number of blocks selected during query-aware selection, thereby affecting both the performance and decoding efficiency of \name.
We evaluate \name~on the 8B size model on HELMET, following Appendix~\ref{sec:reproducibility}'s setting, where we set $k=4096$ and $k_q\in \{512, 1024, 1536, 2048\}$ at inference time. 
We sample 20 entries for each task and report the averaged score, followed by the decoding throughput under the 
$EB=16$ and 16K input setting, in Figure~\ref{fig:sbabl}. 



Changing $k_q/k$ trades long-context performance for decoding efficiency.
Figure~\ref{fig:sbabl} shows a clear performance–throughput trade-off as $k_q/k$ varies. Decreasing $k_q/k$ enforces higher locality, which increases decoding throughput but weakens query-aware selection, limiting recall and reducing task performance. Conversely, increasing $k_q/k$ improves performance at the cost of lower throughput. This tunability allows adjusting $k_q/k$ at inference to meet different requirements: use larger $k_q/k$ when quality is prioritized, and smaller $k_q/k$ when higher throughput is needed. 
Moreover, although the model is trained with $k_q/k$, it generalizes well to other ratios at inference time, demonstrating the robustness and generalizability of the hyperparameter $k_q/k$.
To balance long-context performance and efficiency, we set $k_q/k = 0.25$ in our main experiments, under which \name\ outperforms the baselines in both performance and efficiency.


\begin{table*}[t]
\small

\centering
\setlength{\tabcolsep}{1.8pt} 
\scalebox{0.74}{
\begin{tabular}{l|c|cccc>{\columncolor{pink!20}}c!{\vrule width \heavyrulewidth}c|ccccc>{\columncolor{pink!20}}c}
\toprule
\multirow{2}{*}{Tasks} & \multicolumn{6}{c!{\vrule width \heavyrulewidth}}{\textit{Full Prefill}} & \multicolumn{7}{c}{\textit{Sparse Prefill}} \\
 & \textsc{FullAttn} & \textsc{ShadKV}$_{64}$ & \textsc{ShadKV}$_8$ & \textsc{ArkVale} & \textsc{DMA}$^F$ & \textbf{\name}$^F$ & \textsc{InfLLMv2} & \textsc{ShadKV}$_{64}^M$ & \textsc{ShadKV}$_8^M$ & \textsc{InfLLM}$_{128}$ & \textsc{InfLLM}$_{64}$ & \textsc{DMA}$^S$ & \textbf{\name}$^S$ \\
\midrule
\multicolumn{14}{c}{\small\textbf{\textit{Input Length: 32K}}} \\
\midrule
Recall &
56.3$_{\pm1.8}$ & 31.0$_{\pm0.6}$ & 48.8$_{\pm3.4}$ & 41.6$_{\pm2.5}$ & 10.0$_{\pm1.8}$ & \textbf{56.6}$_{\pm4.2}$ & 45.3$_{\pm2.0}$ & 30.4$_{\pm2.7}$ & \textbf{42.6}$_{\pm3.2}$ & 15.0$_{\pm3.2}$ & 15.6$_{\pm1.9}$ & 8.4$_{\pm2.7}$ & 35.0$_{\pm2.5}$
\\
RAG & 
50.7$_{\pm0.5}$ & 49.1$_{\pm1.3}$ & 49.3$_{\pm2.0}$ & 50.2$_{\pm1.5}$ & 47.9$_{\pm2.5}$ & \textbf{51.4}$_{\pm1.4}$ & 49.7$_{\pm2.2}$ & 44.5$_{\pm0.9}$ & \textbf{46.4}$_{\pm1.6}$ & 44.9$_{\pm2.9}$ & 45.1$_{\pm3.1}$ & 32.2$_{\pm3.9}$ & 44.7$_{\pm3.2}$
\\
ICL & 
67.3$_{\pm2.6}$ & \textbf{66.7}$_{\pm5.7}$ & \textbf{66.7}$_{\pm2.5}$ & 65.7$_{\pm4.9}$ & 59.3$_{\pm4.0}$ & 64.7$_{\pm3.8}$ & 69.7$_{\pm2.5}$ & 60.0$_{\pm1.7}$ & 64.0$_{\pm1.4}$ & 38.3$_{\pm2.6}$ & 38.3$_{\pm1.2}$ & 53.3$_{\pm3.3}$ & \textbf{68.7}$_{\pm5.4}$
\\
Cite & 
13.9$_{\pm3.7}$ & 9.9$_{\pm2.3}$ & 10.7$_{\pm0.6}$ & 9.7$_{\pm1.5}$ & 10.3$_{\pm0.2}$ & \textbf{12.7}$_{\pm1.1}$ & 12.6$_{\pm3.7}$ & 9.8$_{\pm1.2}$ & 9.2$_{\pm1.9}$ & 13.0$_{\pm1.7}$ & \textbf{10.9}$_{\pm1.4}$ & 9.5$_{\pm1.3}$ & 10.0$_{\pm1.9}$
\\
Rerank & 
30.8$_{\pm0.5}$ & 9.1$_{\pm0.5}$ & 19.6$_{\pm1.9}$ & 19.4$_{\pm1.9}$ & 13.3$_{\pm1.0}$ & \textbf{22.0}$_{\pm4.0}$ & 20.0$_{\pm3.3}$ & 6.6$_{\pm1.8}$ & 13.8$_{\pm2.2}$ & 4.3$_{\pm1.0}$ & 4.8$_{\pm1.7}$ & 3.2$_{\pm0.8}$ & \textbf{16.4}$_{\pm1.6}$
\\
LongQA & 
38.0$_{\pm7.0}$ & 34.7$_{\pm4.7}$ & 36.3$_{\pm5.6}$ & 32.8$_{\pm3.2}$ & \textbf{34.8}$_{\pm5.7}$ & 34.2$_{\pm6.8}$ & 32.5$_{\pm4.7}$ & 33.2$_{\pm3.2}$ & 32.3$_{\pm2.6}$ & 10.6$_{\pm1.4}$ & 10.9$_{\pm3.4}$ & \textbf{33.9}$_{\pm1.4}$ & 30.6$_{\pm6.0}$
\\
Summ. & 
10.1$_{\pm2.7}$ & \textbf{10.9}$_{\pm2.2}$ & 11.2$_{\pm3.0}$ & 8.2$_{\pm2.0}$ & 4.5$_{\pm1.2}$ & 10.6$_{\pm0.9}$ & 8.3$_{\pm2.1}$ & 8.0$_{\pm0.7}$ & 5.7$_{\pm1.3}$ & 9.0$_{\pm0.8}$ & 7.9$_{\pm2.4}$ & 7.3$_{\pm2.9}$ & \textbf{9.6}$_{\pm3.4}$
\\
\midrule
Avg. $\uparrow$ & 
38.2$_{\pm1.3}$ & 30.2$_{\pm1.4}$ & 34.6$_{\pm1.6}$ & 32.5$_{\pm1.1}$ & 25.7$_{\pm1.5}$ & \textbf{36.0}$_{\pm1.5}$ & 34.0$_{\pm1.9}$ & 27.5$_{\pm0.7}$ & 30.6$_{\pm0.6}$ & 19.3$_{\pm1.2}$ & 19.1$_{\pm0.9}$ & 21.1$_{\pm0.6}$ & \textbf{30.7}$_{\pm2.2}$
\\
\midrule
\multicolumn{14}{c}{\small\textbf{\textit{Input Length: 64K}}} \\
\midrule
Recall &
38.2$_{\pm2.8}$ & 20.5$_{\pm0.8}$ & 26.3$_{\pm2.6}$ & 23.0$_{\pm2.9}$ & 5.4$_{\pm0.9}$ & \textbf{40.4}$_{\pm1.7}$ & 30.2$_{\pm2.6}$ & 16.6$_{\pm1.4}$ & \textbf{29.0}$_{\pm4.6}$ & 12.2$_{\pm1.9}$ & 12.2$_{\pm2.0}$ & 6.8$_{\pm2.0}$ & 27.0$_{\pm1.5}$
\\
RAG & 
44.9$_{\pm2.6}$ & 44.8$_{\pm1.9}$ & 44.7$_{\pm3.2}$ & \textbf{45.3}$_{\pm2.2}$ & 40.5$_{\pm3.5}$ & 45.0$_{\pm3.5}$ & 42.4$_{\pm1.1}$ & 38.6$_{\pm1.4}$ & 38.2$_{\pm1.9}$ & \textbf{42.6}$_{\pm3.6}$ & 41.7$_{\pm3.2}$ & 32.9$_{\pm5.3}$ & 42.5$_{\pm1.5}$
\\
ICL & 
58.0$_{\pm3.7}$ & 56.7$_{\pm3.7}$ & 59.3$_{\pm2.9}$ & \textbf{61.3}$_{\pm3.7}$ & 59.7$_{\pm1.2}$ & 49.3$_{\pm1.2}$ & 70.3$_{\pm4.9}$ & 59.3$_{\pm3.4}$ & 59.7$_{\pm2.6}$ & 44.0$_{\pm5.0}$ & 40.0$_{\pm2.2}$ & 58.0$_{\pm0.8}$ & \textbf{71.3}$_{\pm6.1}$
\\
Cite & 
12.9$_{\pm1.8}$ & 9.0$_{\pm1.5}$ & 10.0$_{\pm2.0}$ & \textbf{12.1}$_{\pm0.7}$ & 11.0$_{\pm1.2}$ & 10.0$_{\pm1.3}$ & 8.8$_{\pm2.0}$ & 7.0$_{\pm1.1}$ & 8.3$_{\pm1.8}$ & \textbf{12.1}$_{\pm1.5}$ & 11.5$_{\pm2.5}$ & 7.0$_{\pm1.6}$ & 7.6$_{\pm0.9}$
\\
Rerank & 
11.2$_{\pm1.7}$ & 0.5$_{\pm0.2}$ & 2.8$_{\pm1.1}$ & \textbf{9.5}$_{\pm1.2}$ & 7.4$_{\pm0.4}$ & 8.6$_{\pm0.7}$ & 6.8$_{\pm2.6}$ & 0.8$_{\pm0.8}$ & 0.8$_{\pm0.7}$ & 1.8$_{\pm0.3}$ & 1.7$_{\pm0.4}$ & 3.3$_{\pm0.6}$ & \textbf{5.3}$_{\pm0.6}$
\\
LongQA & 
38.1$_{\pm6.7}$ & 36.8$_{\pm3.7}$ & \textbf{36.9}$_{\pm4.6}$ & 36.1$_{\pm6.7}$ & 27.2$_{\pm2.0}$ & 36.3$_{\pm3.0}$ & 37.0$_{\pm4.1}$ & 34.3$_{\pm2.2}$ & 34.6$_{\pm2.2}$ & 14.8$_{\pm2.5}$ & 16.9$_{\pm6.8}$ & 29.9$_{\pm4.9}$ & \textbf{35.9}$_{\pm2.4}$
\\
Summ. & 
8.2$_{\pm1.8}$ & \textbf{6.7}$_{\pm2.2}$ & 5.6$_{\pm1.0}$ & 6.2$_{\pm2.5}$ & 3.5$_{\pm0.1}$ & 4.6$_{\pm1.4}$ & 6.3$_{\pm1.3}$ & 2.7$_{\pm0.6}$ & 3.0$_{\pm0.2}$ & \textbf{11.1}$_{\pm3.0}$ & 10.4$_{\pm0.4}$ & 3.9$_{\pm1.3}$ & 5.1$_{\pm2.8}$
\\
\midrule
Avg. $\uparrow$ & 
30.2$_{\pm1.9}$ & 25.0$_{\pm1.7}$ & 26.5$_{\pm1.8}$ & 27.6$_{\pm2.2}$ & 22.1$_{\pm0.9}$ & \textbf{27.7}$_{\pm1.3}$ & 28.8$_{\pm0.8}$ & 22.8$_{\pm0.5}$ & 24.7$_{\pm0.8}$ & 19.8$_{\pm1.5}$ & 19.2$_{\pm1.8}$ & 20.3$_{\pm1.2}$ & \textbf{27.8}$_{\pm1.3}$
\\
\bottomrule
\end{tabular}
}
\caption{\textbf{\textit{Helmet}} scores (\%) of \name~compared with all baselines  on longer input lengths. \name$^F$ uses full attention during the prefill stage and sparse attention during the decode stage, while \name$^S$ adopts sparse attention for both stages.}
\label{tab:results-longer-inputs}
\vspace{-10pt}
\end{table*}

\textbf{Varying the budget $k$.}
To demonstrate that \name~consistently achieves better performance across different levels of attention sparsity (i.e., different budget $k$), we conduct an ablation study on $k$. We benchmark \name~and all baselines on HELMET using the 8B model, setting $k \in {2048, 4096, 6144}$ to vary the sparsity level. We set the sliding window size to 1024 for $k \in {4096, 6144}$ and to 512 for $k = 2048$. For \name, we fix the ratio $k_q / k = 0.25$ across all settings. Other settings are aligned to Appendix~\ref{sec:reproducibility}.
For each task, we sample 20 instances and report the results in Table~\ref{tab:results-budget}.

\textbf{A larger budget improves long-context performance.} As shown in Table~\ref{tab:results-budget}, the average HELMET score increases as the budget $k$ grows. When the sparsity reaches 0.375, \name~shows no performance degradation under either full prefill or sparse prefill compared with \textsc{FullAttn}. In contrast, with smaller budgets, long-context performance decreases accordingly. The results also indicate that {Recall} and {Rerank} are more sensitive to changes in $k$, whereas other tasks, especially {ICL} and {RAG}, remain relatively stable across different budgets.

\textbf{\name~outperforms all baselines across various budgets.} \name~achieves a higher average HELMET score than all other baselines under both full-prefill and sparse-prefill settings. As the budget decreases, \name~maintains strong performance, whereas the baselines exhibit notable performance degradation. For more complex long-context tasks, such as {Cite} and {Rerank}, the baselines suffer a larger degree of degradation compared with \name. These results demonstrate that \name~is more robust to varying attention sparsity.

\subsubsection{Efficiency Under Various Semantics}
\label{sec:sem-eff}

\begin{table}[h]
\small

\centering
\setlength{\tabcolsep}{5pt} 

{
\begin{tabular}{l|cccccc}
\toprule

Task & T1.1 & T1.2 & T2.1 & T2.2 & T3.1 & T3.2 \\
Context Requirement & Full & Partial & Full & Partial & Full & Partial \\
\midrule
\textsc{InfLLMv2}+NOSI, Offload & 561.0 & 576.7 & 590.0 & 563.5 & 568.4 & 550.1 \\
\name+NOSI, No Offload & 220.8 & 220.3 & 215.5 & 215.4 & 218.6 & 211.4 \\
\name+NOSI, Offload & 700.4 & 712.5 & 714.4 & 702.6 & 709.9 & 695.9 \\
\bottomrule
\end{tabular}
}

\caption{
Decoding throughput (tok/s) $\uparrow$. EB: 16, input length: 16K. ``Ctx. Req.'' refer to \textit{LongBench-Pro}'s context requirement.
}
\label{tab:results-semantic-locality}
\end{table}

Since \name~relies on semantic locality to achieve higher decoding throughput, the degree of locality may vary across input contexts. To demonstrate that \name~can consistently improve decoding throughput, we benchmark \name~with offloading on \namei~against the strongest decoding-throughput baselines (\textsc{InfLLMv2}+NOSI with offloading, and \name+\namei~without offloading) on LongBench-Pro~\citep{chen2026longbench}. By design, LongBench-Pro categorizes inputs into full-context and partial-context settings, which exhibit different levels of locality.
As shown in Table~\ref{tab:results-semantic-locality}, \name+\namei~ outperforms the baselines under diverse context requirements. Its decoding throughput remains stable in both full-context and partial-context settings, demonstrating the generalizability of the proposed method, \name.





\subsubsection{Various Hardware Configurations}
\label{sec:hardware}

\paragraph{Various NUMA configurations.} 

\begin{table}[h]
\small

\centering
\setlength{\tabcolsep}{5pt} 

{
\begin{tabular}{ccc|c|l}
\toprule
\multirow{2}{*}{\#Core} & \multicolumn{2}{c|}{NUMA Index} & \multirow{2}{*}{Thru.} & \multirow{2}{*}{Comment} \\
& Mem. & Core & & \\
\midrule
26 & 0 & 0 & 312.39 & Main experiment. \\
26 & 0 & 1 & 308.49 & Cross NUMA (complete). \\
26 & 0 & Half 0, Half 1 & 309.22 & Cross NUMA (half). \\
13 & 0 & 0 & 295.86 & 104 cores, 8 GPUs. \\
24 & 0 & 0 & 308.13 & 192 cores, 8 GPUs. \\
\bottomrule
\end{tabular}
}
\caption{
Decoding throughput (tok/s) $\uparrow$ under different NUMA setting. EB: 16, input length: 16K. 
}
\label{tab:results-numa}
\end{table}

We benchmark \name+\namei~under different NUMA settings by varying the number of cores, memory and core assignments, and the number of GPUs, as shown in Table~\ref{tab:results-numa}. The results show that NUMA configuration has little effect on the efficiency of \name+\namei. When reducing the number of cores per GPU, we observe only minor degradation in decoding throughput, and 24/26 cores already achieve strong throughput. Notably, this configuration can be easily supported by a 192-core or 208-core machine, showing that \name+\namei~remains effective under various hardware configurations.

\paragraph{Efficiency on Improved Hardware.}

\begin{table}[h]
\small

\centering
\setlength{\tabcolsep}{3.0pt} 

{
\begin{tabular}{l|cccc}
\toprule

Input Length & \makecell{InfLLMv2 + NOSI\\ Offload} & \makecell{NOSA+NOSI \\No Offload} & \makecell{NOSA+NOSI \\Offload} \\
\midrule
16K & 1762.16 & 219.38 & \textbf{2551.66} \\
32K & 1535.70 & 436.45 & \textbf{2024.04} \\
64K & 1510.80 & OOM & \textbf{2034.64} \\
\bottomrule
\end{tabular}
}

\caption{
Decoding throughput (tok/s) $\uparrow$ on NVIDIA H100 and PCIe 5.0. Effective batch size: 16.
}
\label{tab:better-hardware}
\end{table}

To investigate whether \name+\namei~can maintain strong efficiency on more advanced hardware, we benchmark the proposed method on NVIDIA H100 GPUs with PCIe 5.0. As shown in Table~\ref{tab:better-hardware}, all methods achieve higher decoding efficiency due to the increased bandwidth of both HBM and the PCIe bus. Nevertheless, \name+\namei~still outperforms the baselines and achieves significantly higher throughput.

\paragraph{More Discussion.}

Communication media can also be NVLink (which is significantly faster than IB), or disk bus (significantly slower). Here, we would like to provide a short discussion on these scenarios.
For NVLink, it is rare to have a cluster where the GPU and CPU are connected via NVLink. For GPU-GPU connections via NVLink, the other GPU is commonly used as a compute node rather than a KV-cache buffer. Therefore, we believe factoring out NVLink as the GPU-buffer transmitter is a reasonable choice.
For slower memory systems, such as the I/O bus of a disk, \name~with\namei can potentially work on these scenarios due to enhanced locality. Due to resource limit, we would not be able to test them now, and we leave them to future work.

\subsubsection{Wall-Time Breakdown Analysis}

\begin{figure*}[t]
\begin{center}
\includegraphics[width=\linewidth]{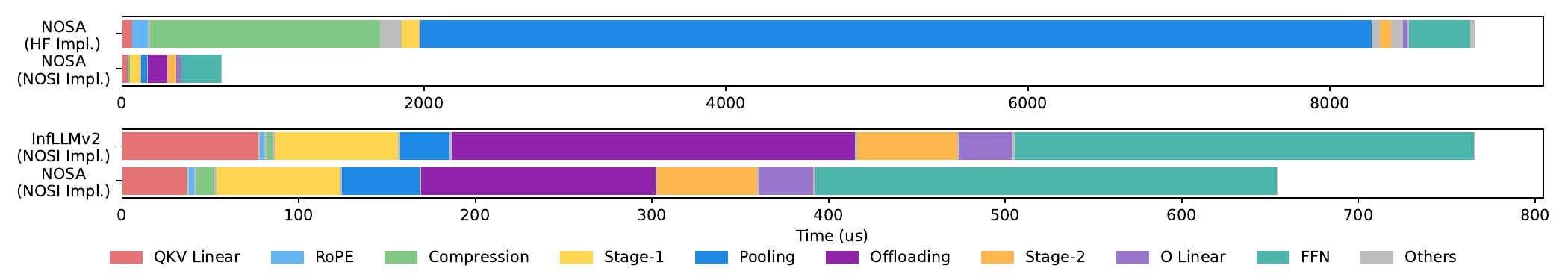}
\end{center}
\vspace{-5pt}
\caption{Decoding wall-time breakdown analysis.
}
\vspace{-10pt}
\label{fig:breakdown}
\end{figure*}

To demonstrate the necessity of constructing the \namei~framework rather than evaluating efficiency within the Hugging Face (HF) implementation, and to highlight the impact of reduced KV-loading time, we conduct a wall-time breakdown analysis on the 8B model with a 16K input length and an equivalent batch size of 16. We compare \name~with \namei~implementations against the Hugging Face implementation, and additionally include a comparison with \textsc{InfLLMv2} implemented in \namei. The results are shown in Figure~\ref{fig:breakdown}. We decompose the decoding time of a single Transformer layer into nine components, and group all remaining overheads (e.g., kernel launches, tensor reshaping, and CPU-side processing) under “Others”.

\textbf{The vanilla Hugging Face implementation is highly inefficient and fails to expose the KV-loading overhead.} As shown in the upper half of Figure~\ref{fig:breakdown}, the HF implementation of \name~incurs extremely large latencies in block compression and pooling, mainly due to excessive \texttt{PyTorch} kernel invocations and inefficient execution. Under this implementation, the KV-loading time in \name~(the purple segment in the second line) is negligible compared with the time spent on compression and pooling. Consequently, optimizing KV-loading yields little end-to-end benefit when using the HF implementation. In contrast, \namei~provides a much more efficient implementation, which exposes KV-loading as a non-trivial component of the decoding latency. Therefore, \namei~enables a meaningful evaluation of KV-loading optimizations and can demonstrate their practical impact on end-to-end performance.

\textbf{\name~achieves a significant reduction in KV-loading time compared with \textsc{InfLLMv2}.} By explicitly enforcing a locality constraint, \name~loads fewer blocks from CPU at each decoding step, resulting in substantially lower KV-loading latency than \textsc{InfLLMv2}, as shown in the lower half of Figure~\ref{fig:breakdown}. In particular, the KV-loading time is reduced by about half. In addition, \name~reduces the QKV linear projection time by using a fused kernel that splits the QKV tensor while simultaneously computing the importance scores. In contrast, \textsc{InfLLMv2} does not compute importance scores and relies on \texttt{PyTorch}'s \texttt{split} operator, which incurs higher overhead than our fused kernel.

\subsubsection{Case Study}
\label{sec:case-study}

\begin{wrapfigure}{r}{0.5\textwidth}  
  \centering
  \vspace{-10pt}
  \includegraphics[width=\linewidth]{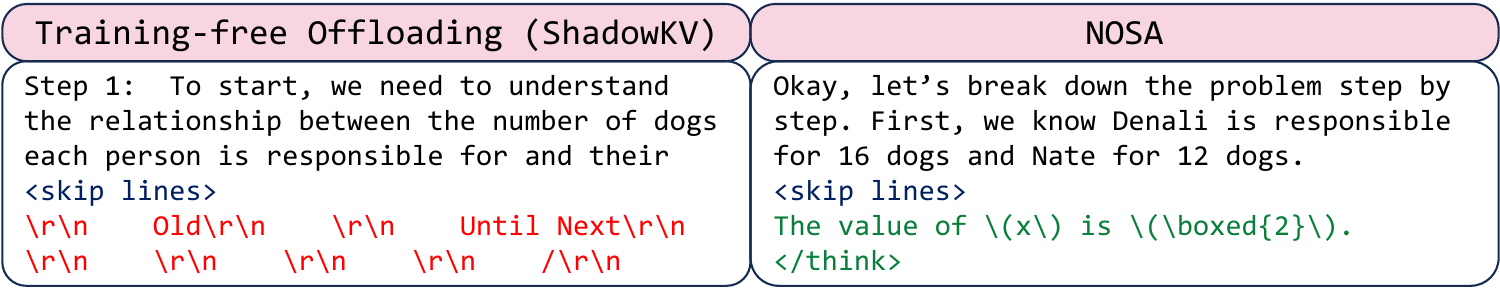}
  \vspace{-10pt}
  \caption{
    Case study of a generation on MATH-500.
  }
  \vspace{-10pt}
  \label{fig:case}
\end{wrapfigure}

We provide a case study in Figure~\ref{fig:ppl} and Figure~\ref{fig:case}. As generation length increases, training-free offloading methods (e.g., \textsc{ShadowKV}) can suffer from a training-inference mismatch, leading to accumulated approximation errors over generation. As a result, the generated text becomes corrupted, as shown in Figure~\ref{fig:case}, whereas \name~maintains generation quality.

To verify this quantitatively, we sample two generations from \textsc{ShadowKV} and \name, and report perplexity throughout the generation process. Perplexity is computed on prefixes from the input up to each position. For a fair comparison, we use the \textsc{FullAttn} 8B model as the evaluator to score both generations. As shown in Figure~\ref{fig:ppl}, \textsc{ShadowKV} exhibits a sharp increase (i.e., an explosion) in perplexity, while \name's perplexity remains consistently low. Higher perplexity indicates increased uncertainty in next-token prediction, which typically leads to more random generations and degraded text quality, matching the observed failure cases of \textsc{ShadowKV}. In contrast, this behavior is not observed for \name, suggesting that maintaining consistency between training and inference prevents error accumulation and enables consistently high-quality long-generation.

\section{Related Works}

Here, we briefly review sparse attention and offloading.
Other related methods are discussed in Appendix~\ref{sec:baselines}.

\textbf{Sparse Attention.} Sparse and approximate attention mechanisms have been proposed attention's dense computation and memory access. 
Early approaches~\citep{xiao2023efficient, han2023lm} employ simple and fixed sparsity patterns. 
More advanced methods~\citep{ge2023model,jiang2024minference,xu2025xattention,xiao2024duoattention,zhang2024sageattention2,zhang2025sageattention2++} introduce richer or more dynamic sparsity structures for improved task performance, often combined with other optimizations like low-bit quantization~\citep{zhang2025sageattention3,zhang2025sageattention,yang2024post}. 
Among these methods, query-aware sparsification is a widely adopted strategy~\citep{li2024snapkv,zhang2023h2o,liu2023scissorhands}. 
In contrast, query-agnostic methods~\citep{huang2024locret,kim2024infinipot,yao2024sirllm} focus on selecting essential KV pairs without queries, but suffer from larger performance degradation.

Recent studies~\citep{deepseekai2024deepseekv32, shi2025trainable, yan2026fsa, vasylenko2025long, gonccalves2025adasplash} focus on eliminating the mismatch between training and inference when applying sparse attentions, i.e., integrating sparsity during model training.
\textsc{NSA}-like methods~\citep{yuan2025native, gao2024seerattention, lu2025moba} introduce trainable block-sparse patterns tailored for long-context scenarios, though these approaches tend to slow down short-context inference. 
\textsc{InfLLMv2}~\citep{zhao2025infllm} addresses this limitation by designing dense-sparse switchable patterns for short and long contexts. 
Beyond LLMs, sparse attentions have also been applied to various domains~\citep{zhang2025vsa,wang2025sparse, yoshai2025hierarchical}.


Note that we omit some previous related method, e.g., \textsc{MoBA}, \textsc{NSA}, and \textsc{SeerAttn} in our comparison, since they are proven to be on-par or weaker than our selected baselines. \textsc{InfLLMv2} is a direct successor to NSA and has been shown to outperform NSA. SSA~\citep{shen2026ssa} provides experimental evidence that MoBA is substantially weaker than NSA. \textsc{SeerAttn} is weaker than \textsc{Quest}~\citep{tang2024quest} in, which is in turn weaker than \textsc{ShadowKV}~\citep{sun2024shadowkv}.

\textbf{LLM Offloading Systems.} Offloading is a widely adopted technique for LLM inference in memory-constrained environments. 
Early approaches~\citep{sheng2023flexgen,song2024powerinfer} focus on parameter offloading to inference larger LLMs on smaller GPUs.
Block-wise KV offloading strategies are proposed for handling longer input sequences~\citep{tang2024quest, xiao2024infllm, chen2024arkvale} and increasing decoding throughputs~\citep{sun2024shadowkv, chen2024magicpig}.
More recent studies~\citep{yang2025lserve,zhou2025sparseserve,chen2025ess, liu2025clusterkv, wu2025louiskv} focusing on supporting large-scale serving systems with sparse attentions to enhance their decoding efficiency.

\section{Conclusion}


We propose \name, a trainable sparse attention mechanism for KV-cache offloading with an explicit locality constraint, and implement an inference system \namei~to show its efficiency.
We evaluate performance and decoding efficiency on 1B, 3B, and 8B models.
\name~outperforms KV cache offloading baselines on general, long-context input, and long-generation tasks, while improving decoding throughput by up to $5.04\times$, $1.92\times$, and $1.83\times$ over \textsc{FullAttn}, \textsc{InfLLMv2}, and \textsc{ShadowKV}.
Future work includes integrating \name~and \namei~into LLM serving engines and accelerating the rollout in LLM reinforcement learning.

\section*{Impact Statement}

This paper aims to advance the field of machine learning. We expect our method to improve LLM decoding efficiency, which can lower the compute and energy cost of inference. By making long-context decoding more efficient, our approach may facilitate broader deployment of LLM applications while reducing their environmental footprint.
We do not foresee any potential negative ethical or societal impacts arising from this work.

\section*{Limitations}
Due to time and resource constraints, we evaluate \name~only on models ranging from 1B to 8B parameters, and do not integrate \name~into larger LLMs, alternative architectures, or multimodal LLMs at this stage. 
Moreover, all efficiency results are obtained with \namei~rather than existing serving frameworks such as vLLM and SGLang, as integrating \name~into these systems requires substantial engineering effort.
Future work includes integrating \name~and \namei~into mainstream LLM serving engines, accelerating rollout in LLM reinforcement learning, and evaluating \name~on a broader range of model architectures, such as hybrid models.

\section*{Acknowledgments}

This work is supported by the National Key Research and Development Program of China (2024YFB4505603), National Natural Science Foundation of China (No. 62576186), supported by Tsinghua University Initiative Scientific Research Program. Chaojun Xiao is supported by the China National Postdoctoral Program for Innovative Talents (grant no. BX20250388).


\bibliography{main}
\bibliographystyle{tmlr}

\clearpage

\appendix

\section{Pseudocode of \name}
\label{sec:pseudocode}

\begin{algorithm}[h]
    \small
    \caption{\name's attention calculation (\namei~implementation) at the $t$-th decoding step.}
    \label{alg:infer}
    \begin{algorithmic}
        
    \STATE {\bfseries Input: }{Hidden state $\mathbf{H}\in\mathbb{R}^{B\times 1\times d}$, Linear/eviction head weights $\mathbf{W}_{\text{qkv}}, \mathbf{W}_1, \mathbf{W}_2$, Cache engine \textbf{\texttt{engine}}, Hyperparameters \textbf{\texttt{params}} including $k, k_e, k_q$, block size, pooling strides, etc.}
    \STATE {\bfseries Output: }{Attention output $\mathbf{H}'\in\mathbb{R}^{B\times 1\times d}$.}

    \STATE
    \STATE $\mathbf{Q},\mathbf{K},\mathbf{V}, s_t^e = $ \texttt{fused\_linear\_projection}($\mathbf{H}, \mathbf{W}_{\text{qkv}}, \mathbf{W}_1, \mathbf{W}_2$)
    \STATE \texttt{apply\_rope\_inplace}($\mathbf{Q},\mathbf{K}$)
    \STATE
    \STATE $\mathbf{K_c}$ = \textbf{\texttt{cache\_engine}}\texttt{.update\_compressed\_k}($\mathbf{K}$)
    \STATE $\mathbf{S}^e_c, \mathbf{S}^e$ = \textbf{\texttt{cache\_engine}}\texttt{.update\_importance\_score}($s_t^e$)
    \STATE
    \STATE {\bfseries // }{Stage 1: Select Top-$k$ blocks to attend}
    \STATE $\mathbf{S}_c^q = \mathbf{Q}\mathbf{K}_c^\top$
    \STATE \textbf{\texttt{q\_aware\_score}}, \textbf{\texttt{q\_agnostic\_score}} = \texttt{fused\_pooling}($\mathbf{S}_c^q, \mathbf{S}_c^e$, \textbf{\texttt{params}})
    \STATE
    \STATE {\bfseries // }{Select up to $k_q$ tokens from \textbf{\texttt{q\_aware\_score}} to maintain a minimum locality of $k_e/k$}
    \STATE \textbf{\texttt{topk\_idx}} = \texttt{select\_topk\_indices}(\textbf{\texttt{q\_aware\_score}}, \textbf{\texttt{q\_agnostic\_score}}, \textbf{\texttt{params.}}$k_e$, \textbf{\texttt{params.}}$k$)
    \STATE
    \STATE {\bfseries // }{Move cache-missed KV blocks from CPU to GPU}
    \STATE \textbf{\texttt{load\_pos}} = \texttt{generate\_load\_positions}(\textbf{\texttt{topk\_idx}}, \textbf{\texttt{engine.current\_mapping}})
    \STATE \textbf{\texttt{engine}}\texttt{.host\_to\_device\_communication}(\textbf{\texttt{load\_pos}})
    \STATE
    \STATE {\bfseries // }{Stage 2: The real attention calculation}
    \STATE $\mathbf{H}'$ = \texttt{flash\_attn\_with\_kvcache}($\mathbf{Q}$, \textbf{\texttt{cache.key}}, \textbf{\texttt{cache.value}})
    
    \STATE{\bfseries return } $\mathbf{H}'$
    \end{algorithmic}
    
\end{algorithm}

\section{Reproducibility Settings}
\label{sec:reproducibility}

\subsection{Environment}

We conduct training on 32 NVIDIA A800 GPUs across four nodes (8 GPUs per node). Intra-node GPU communication leverages third-generation NVLink, while inter-node communication uses HDR InfiniBand. Each node is equipped with Intel® Xeon® Platinum 8470 CPUs and ran CentOS Linux 7 (Core). Each node comprises two NUMA domains with 52 CPU cores in total, and is provisioned with 1 TB of DDR5 memory configured at 4400 MT/s. GPU–CPU communication is conducted over a PCIe 4.0$\times$16 interface, with a peak bandwidth of 31.5 GB/s. All models are trained using the Megatron framework~\citep{shoeybi2019megatron}. For all settings in the inference experiments, we use a single GPU and 26 CPU cores from the NUMA domain closest to the GPU on the same cluster.
All methods except \textsc{ArkVale} use \texttt{bfloat16} precision in both training and inference, while \textsc{ArkVale} uses \texttt{float16} precision since \textsc{ArkVale} does not have \texttt{bfloat16} kernels.

\subsection{Hyperparameters}

\subsubsection{Model Training}
\label{sec:model-training}

We train 1B, 3B, and 8B models under the pipeline illustrated in Figure~\ref{fig:framework} for \textsc{FullAttn}, \textsc{InfLLMv2}, \textsc{DMA}, and the \name~variant, resulting in a total of 12 models. The model architectures and training configurations are summarized in Table~\ref{tab:model-arch}. For the 1B and 3B settings, we adopt the Llama-3 architecture~\citep{grattafiori2024llama}, while we take MiniCPM4-base~\citep{team2025minicpm4} as our 8B base models. We apply LongRoPE~\citep{ding2024longrope} to extend the 4K context length in the pretraining stage to 16K in the long-context continual pretraining stage. All models share the MiniCPM4 tokenizer. We use the AdamW optimizer~\citep{loshchilov2017decoupled} together with the WSD learning rate scheduler~\citep{hu2024minicpm}. Long-context continual pretraining is conducted on the InfLLMv2-data-5B dataset~\citep{openbmb_InfLLM_V2_data_5B_2025}.

\paragraph{Fairness of Competition with Baselines.}

Here, we would like to clarify the fairness of the competition between our method with training-free offloading baselines. The training-free offloading baselines in our paper are not applied directly to the original short-context base model; they are applied after the same long-context CPT stage, because without long-context adaptation these models cannot reliably handle long contexts. Therefore, all compared methods already rely on long-context CPT. NOSA does not add another round of training, nor does it use additional long-context data. It simply replaces the dense-attention CPT used before post-training offloading with an offloading-aware sparse-attention CPT, so that the model is trained under the locality constraint that will be used at inference time. This design is motivated by the following observations:

\textbf{(I)} Long-context CPT is a standard stage for modern LLMs when extending their context length. Whether one later applies offloading in a training-free manner or trains a sparse/offloaded attention mechanism, this CPT stage is still needed. Therefore, NOSA does not add training cost on top of the long-context adaptation pipeline; it changes the attention mechanism used during that stage.

\textbf{(II)} Recent trainable sparse attention methods, such as NSA and InfLLMv2, show that training with sparse attention during CPT (sparse$\rightarrow$CPT) generally outperforms applying sparsity only after dense CPT (CPT$\rightarrow$sparse). NOSA follows the same motivation, but further incorporates the locality requirement imposed by KV offloading.

\textbf{(III)} Offloading introduces a locality constraint that is absent from ordinary sparse attention. We therefore compare three settings: (a) post-training offloading, where CPT is conducted with dense attention and offloading is applied only afterward; (b) directly offloading a trainable sparse attention model, where CPT uses sparse attention but without explicit locality constraints; and (c) NOSA, where CPT uses sparse attention under offloading-aware locality constraints. Empirically, NOSA provides the best tradeoff between performance and decoding efficiency.

In our experiments, baselines such as \textsc{ShadowKV}, \textsc{InfLLM}, and \textsc{ArkVale} are also applied after long-context CPT with full attention, since the base models are initially pretrained with shorter context lengths. Thus, NOSA should be understood as changing the timing and constraints of sparse/offloaded attention within the standard CPT process, rather than adding a new training stage.

\paragraph{Potential Risks of Model and Dataset Artifacts.}

Both MiniCPM4 models and the InfLLMv2-data-5B dataset are released under the \texttt{Apache-2.0} license, and we comply with the license requirements throughout our research. We use these artifacts for research purposes only, which is consistent with their intended use. We acknowledge that large-scale training datasets may inadvertently contain personally identifiable information or offensive content, and InfLLMv2-data-5B may share such risks. In this work, aligned with common practice in related long-context pretraining studies, we directly use the publicly released long-context pretraining dataset without additional preprocessing. Since we do not perform further data collection or modification, we rely on the dataset provider’s preprocessing procedures, including anonymization and safety filtering, to mitigate these risks. The MiniCPM4 models are multilingual LLMs. They are designed for natural language generation and cover a wide range of domains represented in their pretraining data. Similarly, InfLLMv2-data-5B is a multilingual dataset covering multiple domains, primarily intended for long-context modeling. In total, we use approximately 5B tokens from this dataset, as shown in Table~\ref{tab:model-arch}. All entries are used for training. No validation is conducted during training, and model performance is evaluated only after training on downstream benchmarks.

\begin{table*}[t]
\small

\centering
\scalebox{1}{
\begin{tabular}{l|ccc}
\toprule
Model Size & 1B & 3B & 8B\\
\midrule
Architecture & Llama-3 & Llama-3 & MiniCPM4\\
Use $\mu$P~\citep{yang2021tuning} & \ding{55} & \ding{55} & \ding{51}\\
Vocabulary Size & 73448 & 73448 & 73448 \\
Number of Layers & 28 & 32 & 32\\
Hidden Size & 2048 & 2560 & 4096\\
Number of Attention Heads & 16 & 32 & 32\\
Attention Head Dimention & 128 & 128 & 128\\
Number of KV Heads & 2 & 2 & 2\\
FFN Intermediate Size & 6144 & 10240 & 16384 \\
Position Embedding Type & LongRoPE & LongRoPE & LongRoPE \\
RoPE's~\citep{su2024roformer} $\theta$ & 10000 & 10000 & 10000 \\
\midrule
Pretraining Context Length & 4096 & 4096 & 4096 \\
Long-context Continual Pretraining Context Length & 16384 & 16384 & 16384 \\
Optimizer & AdamW & AdamW & AdamW \\
LR-Scheduler & WSD & WSD & WSD \\
Long-context Continual Pretraining Starting LR & 1.50e-5 & 1.50e-5 & 3.00e-4 \\
Long-context Continual Pretraining Ending LR & 1.34e-5 & 8.62e-6 & 2.00e-4 \\
SFT Starting LR & 7.50e-6 & 7.50e-6 & 2.00e-4 \\
SFT Ending LR & 0.00 & 0.00 & 0.00 \\
\midrule
Pretraining Tokens & 1.5T & 3.8T & 8.0T \\
Long-Context Training Tokens & 2.0B & 3.4B & 5.0B \\
SFT Tokens & 0.6B & 1.3B & 1.9B \\
\midrule
Long-context Continual Pretraining GPU Hours & 336 & 883 & 1496 \\
SFT GPU Hours & 78 & 190 & 377 \\
\bottomrule
\end{tabular}
}
\caption{Model architecture and training details.
}
\label{tab:model-arch}
\end{table*}

\subsubsection{Hyperparameters of each method}
\label{sec:method-params}

\textbf{\textsc{InfLLMv2}.}
We adopt most hyperparameters from the official \textsc{InfLLMv2} configuration. The mean-pooling stride $l_s^{\text{mean}}$ is set to 32 and the mean-pooling kernel size $l_k^{\text{mean}}$ is set to 16. We set the block size $n_b$ to 64. For each query token, we attend to $k=4096$ selected tokens, consisting of one block reserved for the attention sink, 16 blocks for the sliding window, and 47 dynamically selected blocks. This configuration is used for both training and inference.

\textbf{\textsc{DMA}.} We use the same configuration as \textsc{InfLLMv2}, except that the 47 dynamically selected blocks are chosen based on the importance scores predicted by the eviction head.

\textbf{\name.} We follow the \textsc{InfLLMv2} configuration and set $k=4096$, $k_q=1024$, and $k_e=3072$. Specifically, 16 blocks are selected via query-aware selection, while the remaining 31 blocks (out of the 47 dynamic blocks) are chosen via query-agnostic selection. Since both the attention sink and the sliding window are inherently query-agnostic, we also count them as part of query-agnostic selection, resulting in a total of 48 query-agnostic blocks. Therefore, $\gamma(t) \geq 75\%$.

\textbf{\textsc{ShadowKV}.} We also select 4096 tokens for sparse attention in \textsc{ShadowKV}, following the same procedure as above. We evaluate four \textsc{ShadowKV} variants on long-context input benchmarks (LongBench and Helmet): \textsc{ShadKV}$_8$, \textsc{ShadKV}$_{64}$, \textsc{ShadKV}$_8^{M}$, and \textsc{ShadKV}$_{64}^{M}$, where the superscript $^{M}$ indicates that \textsc{MInference}~\citep{jiang2024minference} is applied during the prefill stage.
We set the truncation rank in SVD to 40 for all settings.
For \textsc{ShadKV}$_8$ and \textsc{ShadKV}$_8^{M}$, we set the block size to 8, allocate 256 tokens to the sliding window and 768 tokens to outliers, and reserve the remaining 3072 tokens for dynamic selection. For \textsc{ShadKV}$_{64}$ and \textsc{ShadKV}$_{64}^{M}$, we use the same token allocation while increasing the block size to 64. 
The two settings are included because the official \textsc{ShadowKV} configuration uses a block size of 8, and we additionally evaluate a block size of 64 to enable a fair comparison with \name.
For long-generation benchmarks (reasoning tasks), we use the \textsc{ShadKV}$_8$ setting.
Note that the official implementation of \textsc{ShadowKV} supports a maximum generation length of only 1024 tokens, which is incompatible with our setting of generating 8192 tokens. Moreover, the generated tokens are retained in GPU memory and treated as a lengthening sliding window, which leads to an unfair comparison: by the end of an 8K-token generation, \textsc{ShadowKV} activates a 12K context, whereas other methods activate only 4K. To enable long generation for \textsc{ShadowKV} while ensuring a fair comparison under a 4K budget, we implement a V cache offloading strategy and factorize the K cache using SVD bases computed during prefill for long-generation tasks. This allows \textsc{ShadowKV} to operate within the same 4K context budget as the other methods.
For the efficiency benchmark, we cannot use 3072 dynamic tokens because the official \textsc{ShadowKV} offloading kernels do not support this setting (we run in simulation mode for performance evaluation, following the official setup).  Therefore, we set the number of dynamic tokens to 2048, the sliding-window size to 1280, and keep 768 tokens for outliers. Note that this configuration favors \textsc{ShadowKV}'s decoding efficiency and disadvantages \name. Nevertheless, \name~remains faster than \textsc{ShadowKV} under this setting, and thus outperforms \textsc{ShadowKV} in terms of efficiency.

\textbf{\textsc{ArkVale}.} We also select 4096 tokens for sparse attention in \textsc{ArkVale}. The block size is set to 32, the attention sink size is set to 64, and the sliding window size is set to 1024 to ensure a fair comparison. Note that the official implementation of \textsc{ArkVale} does not support \texttt{bfloat16} inference; therefore, we use \texttt{float16} instead. A block size of 64 is also not supported, so we use 32. This configuration favors \textsc{ArkVale}'s performance. Nevertheless, since \name~achieves higher accuracy on the benchmarks than \textsc{ArkVale}, it still outperforms \textsc{ArkVale}. Note that \textsc{ArkVale} does not support the GQA head ratio of 16:1 (query:key) in its official implementation, and thus falls back to MHA.

\textbf{\textsc{InfLLM}.} We retain 4096 tokens in \textsc{InfLLM}. We evaluate two variants, \textsc{InfLLM}$_{64}$ and \textsc{InfLLM}$_{128}$, with block sizes set to 64 and 128, respectively. Since the original \textsc{InfLLM} design uses a block size of 128, we additionally include the 64-block setting for a fair comparison with \name. We set the sliding-window size to 1024 and reserve one block for the attention sink. We set the representative top-$k$ to 4, the maximum cached blocks to the number of activated blocks, and the execution block size to 512.

\subsection{Datasets}
\label{sec:datasets}

For long-context evaluation, we compare \name~with baselines on LongBench~\citep{bai2024longbench} and HELMET~\citep{yen2024helmet}. We use only English tasks with an average input length exceeding 16K in LongBench to ensure higher data quality. HELMET allows controllable context lengths, and we set the input length to 16K. Notably, HELMET's recall subset consists of four challenging retrieval tasks (MK2, MK3, MV from RULER~\citep{hsieh2024ruler}, and JsonKV) which collectively highlight the ability to handle long-context retrieval.
Notably, since some baselines do not apply sparsity during the prefill stage, we report results for full prefill and sparse prefill separately to ensure a fair comparison.
For general tasks, we test on MMLU~\citep{wang2024mmlu}, MMLU-Pro~\citep{wang2024mmlupro}, BBH~\citep{suzgun2022challenging}, GSM8K~\citep{cobbe2021training}, MATH~\citep{hendrycksmath2021}, DROP~\citep{dua2019drop}, MBPP~\citep{austin2021program}, and HumanEval~\citep{chen2021codex}.
For reasoning tasks, we report results on MATH-500~\citep{lightman2023lets} and the Gaokao benchmarks~\citep{zhang2023evaluating}, including math (Gaokao-MS and Gaokao-MH for the science and humanities subsets) and physics (Gaokao-Phy). Since the models are not trained on reasoning traces, we employ a 2-shot prompt (22K tokens) to elicit the model's reasoning ability. 
We evaluate the decoding efficiency on PG19~\citep{raecompressive2019} dataset.

\textbf{Task Abbreviations.} Due to space limitations, we use abbreviations in Table~\ref{tab:results-longbench} and Table~\ref{tab:results-general}. We provide the mapping between task names and their corresponding abbreviations below. Task abbreviations of \textit{LongBench}: Gov\_Report (GO), Triviaqa (TQ), NarrativeQA (NQ), QMSum (QS), Musique (MU), 2wikimqa (2W), MultifiedQA\_En (MQ), RepoBench-P (RB), HotpotQA (HQ), Trec (TR), Passage\_Retrieval\_En (PR), Passage\_Count (PC), and SamSum (SA). 
Task abbreviations of \textit{general tasks}: MMLU (MM), MMLU-Pro (MMP), BBH (BB), GSM8K (GS), MATH (MA), DROP (DR), MBPP (MB), and HumanEval (HE). For general tasks with short sequences, sparsity and attention bias are not applied.

\subsection{Benchmark Settings}
\label{sec:benchmark-settings}

\textbf{Long-Context Inputs.} We follow the official LongBench and HELMET protocols to benchmark all methods. For LongBench, we use greedy decoding for every method and report per-task scores. For HELMET, the official implementation samples evaluation instances from a large data pool, with a random seed controlling the sampled subset. Same as the official setting, the generation of LongQA and Summ. task in HELMET is evaluated by \texttt{gpt-4o}~\citep{hurst2024gpt}. Following the official protocol, we report the mean score and standard deviation for each task. We also use greedy decoding on HELMET.

\textbf{General Tasks.} We use LM-Evaluation-Harness~\citep{eval-harness} to evaluate general-purpose benchmarks. We report 5-shot results for MMLU and MMLU-Pro, 3-shot for BBH, 5-shot flexible-accuracy for GSM8K, 4-shot for MATH, 0-shot F1 for DROP, 3-shot for MBPP, and 0-shot results on the instruction version of HumanEval. To improve performance, we apply a chat template for BBH, GSM8K, DROP, and HumanEval, but do not apply it for other tasks.

\textbf{Reasoning Tasks.} MiniCPM4-base is not trained on reasoning-specific data, and neither the long-context continual pretraining corpus nor the SFT data contains reasoning traces. As a result, the 8B model does not exhibit reasoning behavior by default. To elicit such behavior, we use a two-shot prompt that demonstrates step-by-step reasoning via in-context learning. The prompt is approximately 22K tokens long, and we allow the model to generate up to 8K tokens. The full prompt is shown in Listing~\ref{lst:prompt}. 
We conduct inference with a temperature of 1.0. Since the Gaokao-Phy dataset is relatively small and exhibits high variance, we repeat the experiments four times and report the average score.
We use \texttt{gpt-4o-2024-11-20}~\citep{hurst2024gpt} as our LLM judge and run the prompt shown in Listing~\ref{lst:judge}. We use greedy generation for the LLM judger.
We do not evaluate the 1B and 3B models as their reasoning performance is weaker.

{
\begin{lstlisting}[language=prompt, caption={Test prompt used in reasoning tasks. Blue-highlighted text denotes comments and is excluded from the actual model input.}, label={lst:prompt}]
Question: ## Task B-1.3. <<[The first example]>>

A ship traveling along a river has covered $24 \mathrm{~km}$ upstream and $28 \mathrm{~km}$ downstream. For this journey, it took half an hour less than for traveling $30 \mathrm{~km}$ upstream and $21 \mathrm{~km}$ downstream, or half an hour more than for traveling $15 \mathrm{~km}$ upstream and $42 \mathrm{~km}$ downstream, assuming that both the ship and the river move uniformly.

Determine the speed of the ship in still water and the speed of the river.



Answer: <think>
Okay, so I need to find the speed of the ship in still water and the speed of the river. 
...
</think>
Let \\( v \\) be the speed of the ship in still water (in km/h) and \\( r \\) be the speed
...
The speed of the ship in still water is \\(\\boxed{10}\\) km/h and the speed of the river



Question: Prove that number $1$ can be represented as a sum of a finite number $n$ of real numbers, less than $1,$ not necessarily  distinct, which contain in their decimal representation only the digits $0$ and/or $7.$ Which is the least possible number $n$? <<[The second example]>>



Answer: <think>
Okay, so the problem is to prove that the number 1 can be represented as a sum of a finite
...
</think>
To prove that the number 1 can be represented as a sum of a finite number \\( n \\) of
...
Thus, the least possible number \\( n \\) is \\(\\boxed{8}\\).



Question: {Input Question} <<[Real input question]>>
\end{lstlisting}
}

\begin{lstlisting}[language=prompt, caption={LLM-judge prompt.}, label={lst:judge}]
You are a judge. Determine whether the generation explicitly contains the answer.
Only explicit mention counts -- implicit reasoning, hints, or derivations do not count.
The wording does not need to match exactly; judge based on semantic equivalence.
Ignore correctness of reasoning; only check if the answer is stated anywhere in the generation.
If the answer appears, even if surrounded by extra text, output "yes".
Otherwise output "no".
Output only "yes" or "no".

Generation: {generation}
Answer: {answer}
\end{lstlisting}

\textbf{Effiency Benchmark.} 
All results in Table~\ref{tab:results-thru} are evaluated on the 8B model.
We introduce \textit{equivalent batch size} (EB) to control the on-GPU KV cache size.
For a method $x$ with per-batch on-GPU KV cache size $M_x$, its equivalent batch size under a real batch size $B$ is defined as
\begin{equation}
    EB_x = \frac{B M_x}{M_{\text{NOSA}}}.
\end{equation}
Conversely, given a target equivalent batch size $EB$, the corresponding real batch size is
\begin{equation}
    B_x = EB \cdot \frac{M_{\text{NOSA}}}{M_x}.
\end{equation}
For example, if $x$ is \textsc{FullAttn}, at a 16K input length, \name~retains only 4K tokens on GPU, yielding $M_{\text{NOSA}}/M_x = 0.25$.
Thus, for \textsc{FullAttn}, $B_x = EB/4$, whereas for \name, $B_{\text{NOSA}} = EB$.
By controlling $EB$, we ensure matched on-GPU KV cache sizes across methods for fair comparison.
For each setting, we report the decoding throughput averaged acrossed 4 runs after 1 warmup run, where the time is measured for decoding 4 tokens at each run. 
We assume that each baseline has been optimized to its fullest extent by the original authors, as efficiency is also a key concern in the corresponding works. Moreover, it would be difficult, and in some cases impossible, to implement and compare all methods within a unified system, since system designs in the KV cache offloading literature are often tailored to specific algorithmic choices.

\section{Baselines}
\label{sec:baselines}

In this section, we analyze how \name~differs from reference and baseline methods, and briefly discuss other potentially related methods.

\subsection{Reference and Baseline Methods}

\textbf{\textsc{InfLLMv2}.}
\textsc{InfLLMv2} is a trainable sparse attention mechanism closely related to \textsc{NSA}~\citep{yuan2025native}, \textsc{MoBA}~\citep{lu2025moba}, and \textsc{SeerAttention}~\citep{gao2024seerattention}. Compared with these methods, it features a simpler design and supports switching between dense and sparse inference modes; its inference code and model checkpoints have also been released. \textsc{InfLLMv2} adopts block-wise sparsity patterns without an explicit locality constraint. That is, when deployed in offloading systems, each query token may fetch an arbitrary number of blocks from CPU to GPU. \name\ improves upon this design by introducing a locality constraint and partitioning the selected blocks into query-aware and query-agnostic components. \textsc{InfLLMv2} further argues that sparse attention should be incorporated into training starting from the long-context continual pretraining stage. We follow this training setting in \name, and show that \name\ achieves notably higher decoding efficiency with no substantial performance degradation compared with \textsc{InfLLMv2}.

\textbf{\textsc{DMA}.} 
\textsc{DMA} is a sparse attention mechanism that uses an importance score computed from each token's key vector to determine which tokens to attend to. It is an element-wise method: instead of selecting blocks, it directly selects individual tokens based on their importance scores. However, because this selection is query-agnostic, it effectively implements an eviction policy, i.e., once a token is discarded, it cannot be recalled. \name\ incorporates query-agnostic selection and implements the eviction head using \textsc{DMA}, since some KV pairs are broadly useful across queries and therefore do not require query-dependent selection. We note that the original \textsc{DMA} paper also proposes a query-aware variant, which corresponds to \textsc{DMA-R} in our experiments. In practice, \textsc{DMA} is ill-suited for recall-heavy tasks, and \textsc{DMA-R} may underperform due to training–inference mismatch. \name\ addresses these limitations by combining query-aware and query-agnostic selection, and by integrating an offloading system that balances performance and efficiency.

\textbf{\textsc{ShadowKV}.} 
\textsc{ShadowKV} is a training-free KV cache offloading system that reduces PCIe communication overhead by applying SVD to the K cache and offloading only the V cache. During prefill, \textsc{ShadowKV} uses full attention and constructs both block representations and a low-rank representation of K. At each decoding step, it first performs query-aware block selection, and then overlaps K reconstruction with fetching V blocks from CPU. Because sparsity is introduced only during decoding, \textsc{ShadowKV} suffers from a non-negligible training–inference mismatch, which can introduce approximation errors in decoding. These errors may accumulate over long generations and lead to noticeable degradation in generation quality. In addition, the overlapped K reconstruction and V fetching is not sufficiently efficient, as it relies on the model's inherent locality, which is substantially weaker than the constrained locality enforced by \name.

\textbf{\textsc{ArkVale}.} The main contribution of \textsc{ArkVale} is its query-aware selection strategy based on a bounding cuboid over the key vectors within each block. This selection is accurate and expressive. However, due to its complexity, this mechanism is difficult to integrate into model training and is therefore used only at inference time in the official \textsc{ArkVale} setting.
Consequently, \textsc{ArkVale} suffers from a training–inference mismatch, which can hurt long-context performance. Moreover, due to implementation overheads, \textsc{ArkVale} is notably slower than other baselines in our experiments and therefore does not improve decoding efficiency.

\textbf{\textsc{InfLLM}.} \textsc{InfLLM} was originally designed to support inference over longer sequences under limited GPU memory. To this end, it combines sparse attention with chunked prefill, enabling sparsity to be applied during the prefill stage. Specifically, it builds block-level representations from the KV cache and computes query-aware relevance scores to select the most relevant blocks for each query token. However, because the sparsity pattern is not learned during training, introducing sparsity into long-context prefill can substantially degrade recall, leading to poor performance on long-context benchmarks. Moreover, the \textsc{InfLLM} system is not well optimized for large batch sizes and therefore fails to achieve high decoding throughput when the batch size increases.

\subsection{Other Potentially Related Methods}

Since \name~is positioned as a KV cache offloading method, we briefly discuss other KV-cache optimizations, which fall into three categories: KV cache compression, quantization, and eviction.

\textbf{KV Cache Compression.}
KV cache compression can also enable larger decoding batch sizes by reducing KV cache memory footprint. Existing KV-cache compression techniques can be broadly categorized into four levels: token-level, head-level, layer-level, and dimension-level.
Token-level compression methods, such as \textsc{ClusterAttn}~\citep{zhang2025clusterattn} and \textsc{SqueezedAttn}~\citep{hooper2025squeezed}, merge adjacent or semantically similar KV units along the sequence length dimension. These approaches can be combined with block-wise sparse attention to enable more accurate KV selection, as in \textsc{ClusterKV}~\citep{liu2025clusterkv}.
Head-level compression alleviates GPU memory pressure by pruning attention heads. 
For example, \textsc{DuoAttn}~\citep{xiao2024duoattention} and \textsc{PruLong}~\citep{bhaskar2025cache} propose post-training procedures to identify heads whose KV states can be removed and prune them accordingly.
Layer-level methods, such as \textsc{LCKV}~\citep{wu2024layer} and \textsc{MiniCache}~\citep{liu2024minicache}, target redundancy across layers by sharing a single set of KV states among multiple layers.
Dimension-level compression reduces the KV representation size via low-rank factorization~\citep{saxena2024eigen, sun2024shadowkv} or by performing attention computation in a latent space~\citep{liu2024deepseek, mu2025sals, singhania2024loki}, thereby lowering peak GPU memory usage.
Most of the above approaches, except for trainable latent-space attention such as \textsc{MLA}~\citep{liu2024deepseek}, introduce a training–inference mismatch, which can substantially degrade long-form generation quality. \textbf{These techniques are orthogonal to \name, since they mainly target the KV cache, that is not modified by \name.} In addition, \textsc{MLA} typically requires substantially more training effort (e.g., training from scratch or an additional conversion stage to transform GQA models into MLA models~\citep{ji2025towards}) than \name. We leave combining \name~with other KV-cache compression techniques for future work.
KV cache compression is also widely used in more complex application settings, including retrieval-augmented generation~\citep{cao2025sparse}, code generation~\citep{zhang2025anchor}, and multimodal models~\citep{jiang2025purekv}. Task semantics have also been leveraged for KV cache optimization~\citep{he2025task}. \textbf{\name~can be applied to these scenarios and tasks as well, since they do not directly conflict with \name's design principles.}

\textbf{KV Cache Quantization.}
KV cache quantization is a mainstream approach to reducing the KV cache memory footprint, and it can additionally benefit from low-bit kernels to accelerate inference. \textsc{KVQuant}~\citep{hooper2024kvquant} and \textsc{KIVI}~\citep{liu2024kivi} quantize the KV cache to 4 bits and 2 bits, respectively, achieving substantial compression compared with the standard half-precision floating-point representation. More recent work has further pushed the limit toward 1-bit KV cache quantization~\citep{zandieh2025qjl, tao2025asymkv}. These methods are potentially compatible with quantized attention techniques, such as \textsc{SageAttention}~\citep{zhang2025sageattention, zhang2024sageattention2, zhang2025sageattention2++, zhang2025sageattention3}. \textbf{\name~is also orthogonal to KV cache quantization}, and we leave a systematic study of their combination to future work.

\textbf{KV Cache Eviction.}
KV cache eviction methods reduce the memory burden by discarding less important KV units. Early approaches such as \textsc{StreamingLLM}~\citep{xiao2023efficient} and \textsc{LM-Infinite}~\citep{han2024lm} rely on fixed eviction patterns to remove cache entries. Classical methods, including \textsc{SnapKV}~\citep{li2024snapkv}, \textsc{H2O}~\citep{zhang2023h2o}, and \textsc{FastGen}~\citep{ge2023model}, introduce more sophisticated eviction policies, including query-aware selection that leverages the current query as prior information to identify important KV entries. Query-agnostic eviction methods, such as \textsc{InfiniPot}~\citep{kim2024infinipot} and \textsc{Locret}~\citep{huang2024locret}, further enable eviction in settings where the query is not immediately available. In general, these methods suffers from irreversible information loss: once a token is evicted, it cannot be retrieved. \textbf{Therefore, KV cache eviction is expected to underperform KV cache offloading. However, these methods provide a strong query-agnostic importance estimator, and they can be integrated as the eviction head in \name.}

\section{Ablation on Eviction Head Implementation}
\label{sec:abl-eviction-head}

Here, we ablate the implementation of the eviction head to evaluate the effectiveness of \name's design. 
The eviction head assigns an importance score $s_j^e$ to each token at position $j$, which serves as the metric for selecting query-agnostic tokens in \name. 
There are several possible approaches to implement this mechanism. 
A straightforward approach is to learn the importance score through a lightweight MLP that takes the hidden state of each token as input, similar to the \textit{retaining head} proposed in \textsc{Locret}~\citep{huang2024locret}. 
Alternatively, DMA~\citep{shi2025trainable} employs a single-layer gated projection based on $\mathbf{v}_j$, the value vector at the $j$-th token, to estimate importance. 
In \name, we perform query-agnostic selection using the pre-exponential value from DMA, motivated by our observation that this step is highly sensitive to numerical precision. 
This variant is referred to as {Exp-Delayed DMA} (ED-DMA). 
For comparison, we also include a variant where no explicit attention bias is applied during forward passes, and only its gradient contribution is preserved during backpropagation. 
This simplified version is termed {Simple DMA} (S-DMA).
We list the bias calculation before query-agnostic selection and the attention calculation in Table~\ref{tab:abl-evict-set}.

We first pretrain a 1B-parameter model with an input sequence length of 8K under dense attention, and then perform long-context continual pretraining with \name~using different eviction head implementations, where the sequence length is extended to 16K. We set the total selection budget $k$ to 4096, with 64 attention sink tokens, 1024 sliding window tokens, and $k_q=1024$.
The resulting models are evaluated on RULER~\citep{hsieh2024ruler}. 
As shown in Table~\ref{tab:abl-evict}, ED-DMA achieves the best performance, remaining nearly lossless compared to the original \textsc{InfLLMv2}. 
Other implementations exhibit noticeable performance degradation, particularly on the MK and MV tasks. 
Notably, merely shifting the exponential operator from the bias calculation to the attention computation yields a 5.7\% improvement in accuracy. 
Based on these results, we adopt ED-DMA as the eviction head implementation for \name.

\begin{table*}[h]
\small

\centering
\setlength{\tabcolsep}{3pt} 
\scalebox{1}{
\begin{tabular}{l|c|ccc}
\toprule
Abbreviation & Full Name & Bias Calculation & Attention Calculation\\
\midrule

\textsc{Locret} & \textsc{Locret} & ${b}_j = \sigma(\mathbf{h}_j\mathbf{W}_1)\mathbf{W}_2$ & $\textbf{o}_i = \sum_{j} \frac{\exp(\mathbf{q}_i\mathbf{k}_j^\top+b_j+m_{ij})\mathbf{v}_j}{\sum_{l}\exp(\mathbf{q}_i\mathbf{k}_l^\top+b_l+m_{il})}$ \\[5pt]
DMA& Dynamic Mask Attention & ${b}_j = \exp(\tau(\mathbf{v_j}\mathbf{W_1})\odot\mathbf{W_2})$ & $\textbf{o}_i = \sum_{j}\frac{b_j \exp(\mathbf{q}_i\mathbf{k}_j^\top+m_{ij})\mathbf{v}_j}{\sum_{l}b_l \exp(\mathbf{q}_i\mathbf{k}_l^\top+m_{il})}$\\[5pt]
ED-DMA & Exp-Delayed DMA & ${b}_j = \tau(\mathbf{v_j}\mathbf{W_1})\odot\mathbf{W_2}$ & $\textbf{o}_i = \sum_{j}\frac{\exp(b_j) \exp(\mathbf{q}_i\mathbf{k}_j^\top+m_{ij})\mathbf{v}_j}{\sum_{l}\exp(b_l) \exp(\mathbf{q}_i\mathbf{k}_l^\top+m_{il})}$\\[5pt]
S-DMA & Simple-DMA & ${b}_j = \tau(\mathbf{v_j}\mathbf{W_1})\odot\mathbf{W_2}$ & $\textbf{o}_i =\sum_{j} \frac{\exp(b_j-b_j\text{.detach()}) \exp(\mathbf{q}_i\mathbf{k}_j^\top+m_{ij})\mathbf{v}_j}{\sum_{l}\exp(b_l-b_l\text{.detach()}) \exp(\mathbf{q}_i\mathbf{k}_l^\top+m_{il})}$ \\[5pt]

\bottomrule
\end{tabular}
}
\caption{Details of various implementations of the eviction head.
}
\label{tab:abl-evict-set}
\end{table*}

\begin{table*}[h]
\small

\centering
\setlength{\tabcolsep}{5pt} 
\scalebox{0.9}{
\begin{tabular}{l|ccccccccccccc|c}
\toprule
Implementation & SG1 & SG2 & SG3 & MK1 & MK2 & MK3 & MV & MQ & VT & CWE & FWE & QA1 & QA2 & Avg. $\uparrow$\\
\midrule
\textsc{InfLLM-V2} &
100.0 & 100.0 & 100.0 & 84.0 & 50.0 & 18.0 & 92.5 & 84.5 & 26.0 & 0.8 & 62.7 & 36.0 & 30.0 & 60.3 \\
\midrule
\textsc{Locret} & 
\textbf{100.0} & 98.0 & 86.0 & 64.0 & 40.0 & 4.0 & 77.0 & \textbf{79.5} & 41.6 & 3.6 & 64.7 & 36.0 & 32.0 & 55.9
\\
DMA & 
\textbf{100.0} & \textbf{100.0} & \textbf{98.0} & 70.0 & \textbf{44.0} & 8.0 & 75.5 & 70.0 & 20.8 & 2.0 & 74.7 & 36.0 & 32.0 & 56.2
\\
S-DMA & 
\textbf{100.0} & 96.0 & \textbf{98.0} & 60.0 & 42.0 & 8.0 & 87.0 & 77.0 & \textbf{94.8} & 3.6 & 60.7 & 30.0 & \textbf{34.0} & 60.9 
\\
\rowcolor{pink!20}
ED-DMA & 
\textbf{100.0} & 98.0 & \textbf{98.0} & \textbf{76.0} & 42.0 & \textbf{14.0} & \textbf{90.5} & 78.5 & 62.4 & \textbf{3.8} & \textbf{68.0} & \textbf{40.0} & \textbf{34.0} & \textbf{61.9} 
\\

\bottomrule
\end{tabular}
}
\caption{RULER evaluation results with various implementations of the eviction head.
}
\label{tab:abl-evict}
\end{table*}

\section{Theoretical Guarantee of Locality}
\label{sec:locality_theorem}

\renewcommand{\thetheorem}{\ref{theorem:locality}}
\begin{theorem}
If the selection process of \name~has budget $k=k_q+k_e$, we have $\forall t\in \{2, \cdots, n\},~\gamma(t) \geq \frac{k_e}{k}$.
\end{theorem}
\renewcommand{\thetheorem}{\arabic{theorem}}

\begin{proof}

Such statement equals to $|\Gamma(t-1)\cap \Gamma(t)|\geq k_e$. We only prove the token-wise selection case below for brevity, since the block-wise selection case follows the same reasoning. For simplicity, we do not consider the newly generated token at the $t$-th step, as it always resides on the GPU. Accordingly, we assume that $k$ tokens are selected from $n$ tokens at both the $(t-1)$-th and $t$-th steps.

Denote the block indexes of query-aware selection at step $t-1$ as 
\begin{align}
    \Gamma_q(t-1) = \{a_1, a_2, \cdots, a_{k_q}\},~\text{such that }~ s^q_{a_1} \geq s^q_{a_2} \geq \cdots \geq s^q_{a_{k_q}}.
\end{align}

Denote the indexes of the Top-$k$ largest importance scores as 
\begin{align}
    \text{ArgTop}_k (\{s_i^e\}_{i=1}^n) = \{b_1, b_2, \cdots, b_k\},~\text{such that }~ s^e_{b_1} \geq s^e_{b_2} \geq \cdots \geq s^e_{b_k}.
\end{align}

Here, $\{b_1, \cdots, b_k; a_1, \cdots, a_{k_q}\}$ is a multiset, since some indices in $\{b_1,\cdots,b_k\}$ may also appear in $\{a_1, \cdots, a_{k_q}\}$. According to \name's selection process, we have $\Gamma(t-1)\subseteq \{b_1, \cdots, b_k; a_1, \cdots, a_{k_q}\}$. Denote $p = |\{b_1, \cdots, b_k; a_1, \cdots, a_{k_q}\}| - |\Gamma(t-1)|$, then $1\leq p \leq k_q$.

Now, consider the elements in $\{b_1, \cdots, b_k; a_1, \cdots, a_{k_q}\}$ that are not contained in $\Gamma(t-1)$. Since query-aware selection is prioritized, we have $\{a_1,\cdots,a_{k_q}\}\subset \Gamma(t-1)$. Due to the monotonic construction, the last $p$ elements in $\{b_1, \cdots, b_k\}$ are not included in $\Gamma(t-1)$.

Therefore, we have 
\begin{align}
    b_1, b_2, \cdots, b_{k-p} \in \Gamma(t-1).
\end{align}
Since $0\leq p \leq k_q$, we must have $b_1, b_2, \cdots, b_{k_e}\in \Gamma(t-1)$. 
Note that all the above derivations are independent of $t$. Similarly, we have $b_1, b_2, \cdots, b_{k_e} \in \Gamma(t)$, thereby implying $|\Gamma(t-1)\cap \Gamma(t)|\geq k_e$.

\end{proof}
\end{document}